\documentclass[runningheads]{llncs}

\usepackage{eccv}

\usepackage{eccvabbrv}

\usepackage{graphicx}
\usepackage{booktabs}
\usepackage{multirow}
\usepackage{array}
\usepackage[table]{xcolor}
\usepackage{caption}
\usepackage{wrapfig}
\usepackage{arydshln}

\definecolor{gold}{HTML}{FFF9E5} 
\definecolor{tablegray}{gray}{0.7}

\usepackage[accsupp]{axessibility}  

\usepackage{hyperref}

\usepackage{orcidlink}

\begin{document}

\title{MSPL: Multi-Step Pseudo-Labeling for Open-Vocabulary Object Detection} 

\titlerunning{MSPL}

\author{Hojun Choi\inst{1}\orcidlink{0009-0009-4121-4519} \and
Youngsun Lim\inst{2}\orcidlink{0009-0006-8791-0964} \and
Jaeyo Shin\inst{1}\orcidlink{0009-0005-3931-938X} \and
Hyunjung Shim\inst{1}\orcidlink{0000-0001-6796-1058}}

\authorrunning{H.~Choi et al.}

\institute{KAIST AI, South Korea\\
\email{\{hchoi256,jaeyo\_shin,kateshim\}@kaist.ac.kr} \and
Boston University, USA\\
\email{youngsun@bu.edu}}

\maketitle
\begin{center}
  \centering
  \includegraphics[width=\linewidth]{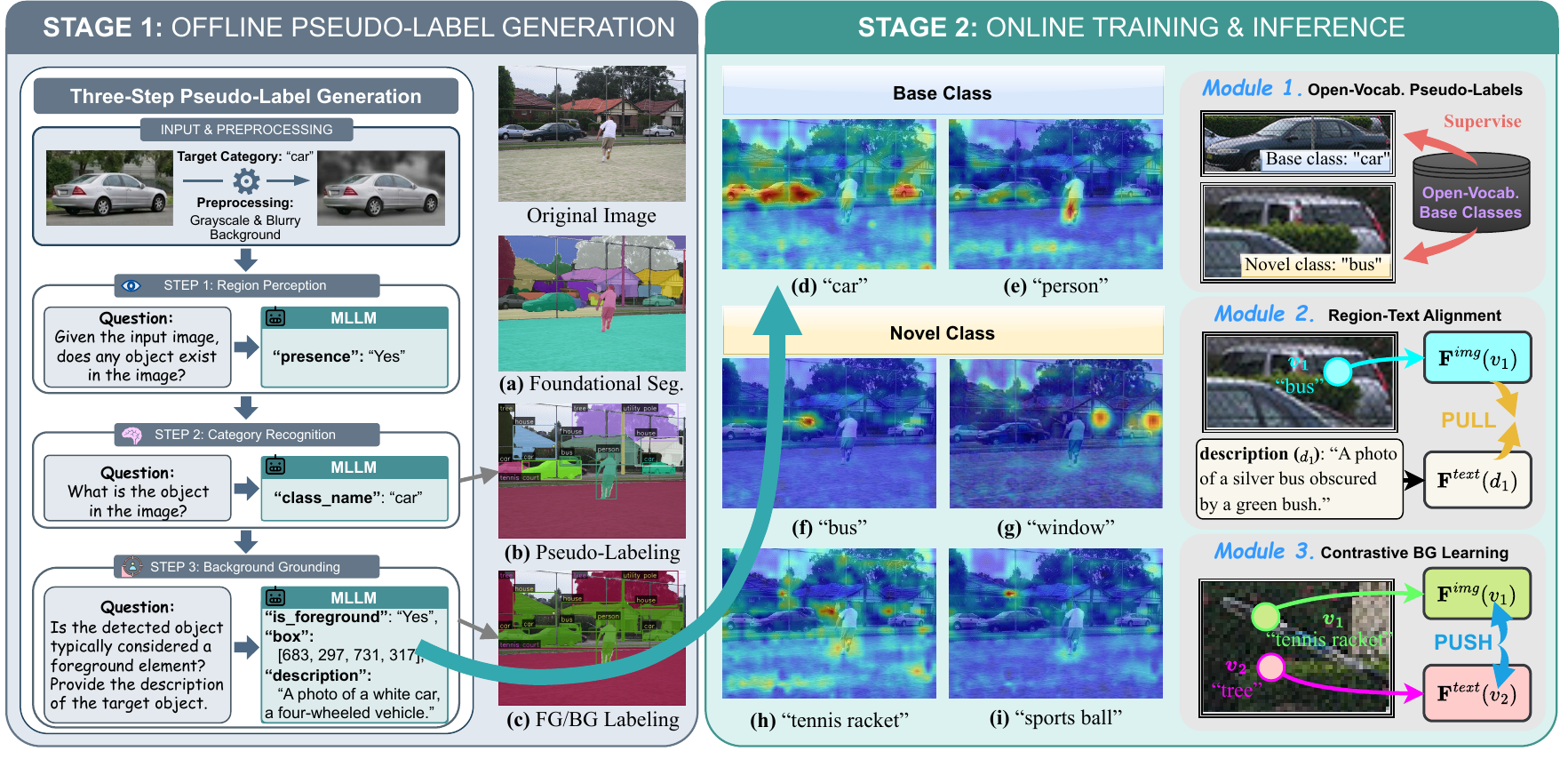}
  \captionof{figure}{\textbf{Offline:} Our method generates robust pseudo-labels via a 3-step prompting process: (a) object localization, (b) category recognition, and (c) background grounding. \textbf{Online:} The open-vocabulary detector leverages these additional supervisory signals to detect both base (d–e) and novel (f) classes, and even unlabeled objects (g–i).}
  \label{fig:teaser}
\end{center}

\begin{abstract}
  Open-vocabulary object detection (OVD) aims to recognize and localize object categories beyond the training set. Recent approaches leverage vision-language models to generate pseudo-labels using image-text alignment, allowing detectors to generalize to unseen classes without explicit supervision. However, these methods depend heavily on single-step image-text matching, neglecting the intermediate reasoning steps crucial for interpreting semantically complex visual contexts, such as crowding or occlusion. In this paper, we introduce MSPL, a framework that incorporates multi-step visual reasoning into the pseudo-labeling process for OVD. It decomposes complex scene understanding into three interpretable steps—object localization, category recognition, and background grounding—where these intermediate reasoning states serve as rich supervision sources. Extensive experiments on standard OVD evaluation protocols demonstrate that MSPL achieves state-of-the-art performance with superior pseudo-labeling efficiency, outperforming the strong baseline by 9.4 AP$_{50}$ for novel classes on OV-COCO and improving box and mask AP$_r$ by 3.2 and 2.2, respectively, on OV-LVIS.
  \keywords{Open Vocabulary \and Object Detection \and Pseudo-Labeling}
\end{abstract}

\section{Introduction}
\label{sec:intro}
Open-vocabulary object detection (OVD) aims to localize both seen (base) and unseen (novel) categories at test time, using only base-class annotations during training. To bridge this supervision gap between seen and unseen categories, recent approaches leverage vision-language models (VLMs) pre-trained on large-scale image-text pairs~\cite{CLIP}. These VLMs map textual descriptions to visual representations, allowing OVD methods to recognize novel classes.

\begin{figure}[tb]
  \centering
  \includegraphics[width=\linewidth]{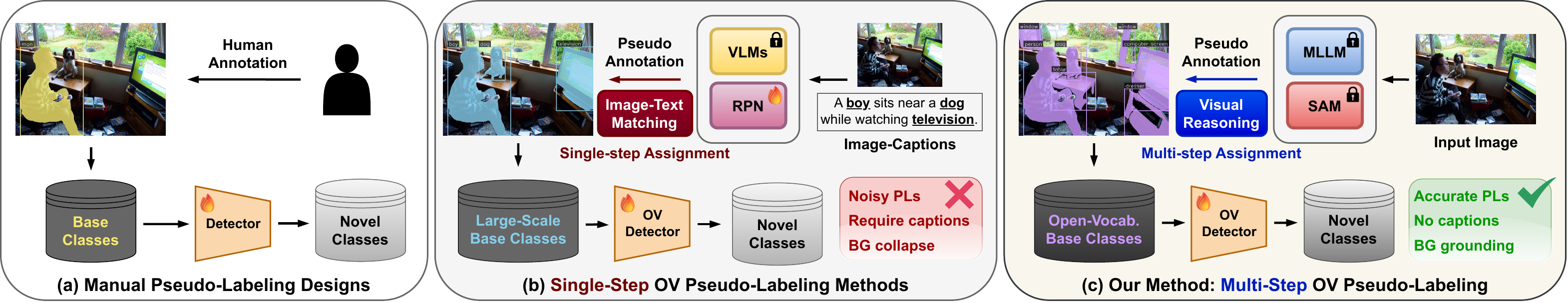}
  \caption{(a) Manual pseudo-labels for novel classes is costly and does not scale. (b) Recent approaches automate this process via single-step semantic assignment with vision-language models and image captions, struggling in complex scenes. (c) Our caption-free method leverages multi-step reasoning to interpret semantically complex scenes.}
  \label{fig:PL_trend}
\end{figure}

Among such efforts, pseudo-labeling has emerged as a state-of-the-art approach for OVD by augmenting the base set with automatically generated annotations that partially cover novel classes~\cite{PB-OVD, LP-OVOD}. In~\cref{fig:PL_trend}-a, early pseudo-labeling methods for OVD relied on manual annotation of novel classes, which was costly and lacked scalability. In~\cref{fig:PL_trend}-b, more recent approaches~\cite{VL-PLM,sasdet} leverage VLMs to automate the generation of pseudo-annotations for novel classes based on the similarity between visual features and text embeddings of potential object categories—including some novel classes—typically derived from image captions~\cite{cococaption}.

\begin{figure}[tb]
  \centering
  \includegraphics[width=\linewidth]{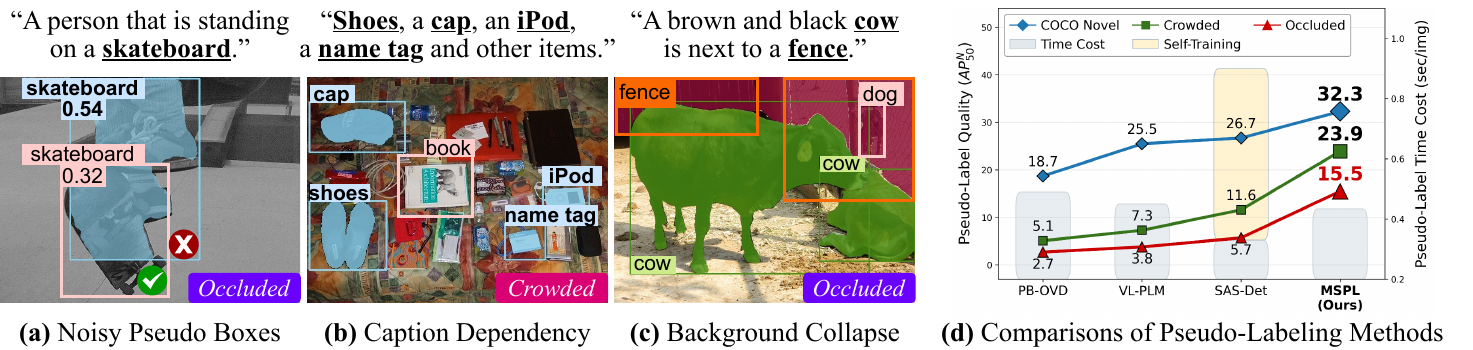}
  \caption{\textbf{Challenges in pseudo-labeling for complex scenes.} (a) Errors in single-step VLM-based semantic assignment, (b) coarse caption semantics, and (c) unlabeled objects treated as background. (d) MSPL generates robust pseudo-labels on ground-truth novel classes in complex scenes at the lowest per-image computational cost.}
  \label{fig:limit}
\end{figure}

Despite their strong performances in general scenes, state-of-the-art OVD approaches still struggle in challenging scenarios involving crowding or occlusion. We identify the root cause as a reliance on single-step image–text matching via CLIP~\cite{clip_drawback}. Because complex scenes require disentangling overlapping visual elements, this direct mapping collapses, leading to three critical failures in pseudo-labeling. \textbf{(L1) Noisy pseudo boxes}: Single-step alignment assigns labels based on surrounding context rather than region-specific content. Since VLMs trained with image-level supervision encode co-occurrence statistics rather than object-level semantics~\cite{RegionCLIP}, a region inherits the label of a contextually dominant neighbor. In~\cref{fig:limit}-a, the crop of partially occluded feet is incorrectly labeled ``skateboard'' due to its strong co-occurrence with the skateboard in the scene. \textbf{(L2) Caption dependency}: Single-step alignment requires a predefined candidate set, making it structurally bound to image captions as the sole category source. Any object absent from the caption or under-described remains undiscovered by design. In~\cref{fig:limit}-b, ``book'' goes entirely unlabeled simply because it is omitted from the caption, while ``iPod'' can be misclassified as a visually similar object (\eg, ``cell phone'') due to its coarse description as a simple class name—a failure inherent to single-step alignment's inability to provide fine-grained discovery beyond the provided candidate set. \textbf{(L3) Background collapse}: Recognizing an occluded object requires sequential reasoning—first identifying the occluder, then inferring the hidden instance. Single-step alignment bypasses this decomposition, causing unmatched region to be erroneously absorbed into the background during training~\cite{lbp}. In~\cref{fig:limit}-c, ``the dog occluded by a fence'' is never assigned any label, and is instead learned as background. This is a direct consequence of single-step reasoning's inability to decompose the scene.

We argue that these limitations stem from a common bottleneck: single-step alignment compresses the entire scene into a single reasoning unit, leaving no room to disentangle co-occurring objects, discover underspecified categories, or reason through occlusion. To overcome this, we propose reformulating pseudo-labeling as an interpretable multi-step visual prompting process \cite{vchatgpt}. Our three-step prompting framework directly addresses all three limitations within a single structured reasoning pass: (1) \textit{object localization} grounds each region in object-level visual evidence via SAM~\cite{SAM}, bypassing co-occurrence bias (L1); (2) \textit{category recognition} assigns zero-shot labels and their region description via MLLM reasoning without caption-derived candidates, enabling fine-grained discovery of any object in the scene (L2); and (3) \textit{background grounding} explicitly identifies background concepts to disentangle them from occluded foreground instances (L3). In the online phase, a detector is trained under a contrastive objective using the resulting pseudo-labels along with their intermediate reasoning outputs. By shifting complex reasoning offline, this decoupled strategy minimizes training overhead and structurally isolates noisy supervision from the training gradient, ensuring only high-fidelity annotations propagate into online learning.

Importantly, this design is not a naive combination of existing models. In fact, directly integrating SAM and MLLMs without structured reasoning fails on two fronts: SAM's class-agnostic masks span inconsistent semantic granularities, incurring redundant MLLM inference to hallucinate labels for partial regions; and single-pass MLLM queries not only degrade label accuracy but eliminate the intermediate reasoning states providing essential supervision for online training. Instead, our principled multi-step design addresses the root causes of single-step failures (L1–L3), enabling SAM and MLLM to work synergistically to generate high-fidelity, exclusively object-level annotations where a direct integration would otherwise be computationally prohibitive and prone to collapse.

We conduct extensive experiments on two OVD benchmarks, OV-COCO~\cite{coco} and OV-LVIS~\cite{lvis}. In~\cref{fig:limit}-d, under two challenging conditions such as crowding and occlusion, our method demonstrates superior pseudo-label quality compared to previous pseudo-labeling methods~\cite{PB-OVD,VL-PLM,sasdet}, with the most competitive runtime. Furthermore, our method sets a new state-of-the-art, improving box $AP_{50}$ for novel classes on OV-COCO by 9.4, and further enhancing both box and mask $AP_r$ on OV-LVIS by 3.2 and 2.2 respectively, compared to prior work~\cite{baron}.

\section{Related Work}
\label{sec:related}
\subsection{Multi-Step Reasoning in Vision-Language Models}
Multi-step prompting addresses complex tasks by decomposing them into intermediate, interpretable queries. In language models, such intermediate reasoning improves performance on complex reasoning problems~\cite{cot}. This idea has been extended to vision and vision-language tasks through structured prompts or predictions, including bounding boxes~\cite{visualcot}, future states for autonomous driving~\cite{drivevlm}, image infillments~\cite{cotclip}, and synthesized CLIP embeddings~\cite{cotdiffusion}. It has also been used in embodied tasks such as textual planning~\cite{embodiedgpt}, reward guidance~\cite{zhang}, and robotic sub-goal generation~\cite{cotvla}. In this work, we apply multi-step prompting to open-vocabulary object detection by decomposing pseudo-label generation into interpretable stages for robust labeling in complex scenes.

\subsection{Open-Vocabulary Object Detection}
Open-vocabulary object detection (OVD) aims to detect novel objects unseen during training by leveraging vision-language models (VLMs)~\cite{CLIP} trained on large-scale image-text pairs. Recent OVD methods~\cite{detpro,cora,promptdet} use prompt modeling to transfer knowledge through learned prompts, while others align detectors with VLM features via knowledge distillation~\cite{vild,baron}. Some approaches enhance the text modality with large language models~\cite{dvdet,shine}, improve novel class prediction using synthetic images~\cite{instagen}, or facilitate cross-modal information exchange for prompt-based detection~\cite{groundingdino}. Another line of work~\cite{rovit,fvlm} fine-tunes VLMs with learnable parameters for feature extraction, which is often computationally costly.
Recently, pseudo-labeling methods~\cite{PB-OVD,VL-PLM,sasdet} have used caption-derived supervision~\cite{cococaption} to expand beyond restricted base classes. However, their vocabulary coverage depends on caption quality and content, while requiring additional annotations. In contrast, our caption-free approach leverages the zero-shot capabilities of powerful MLLMs to expand the vocabulary without costly caption annotations. Moreover, to overcome the limited reasoning ability of direct, single-step CLIP matching in complex scenes~\cite{clip_drawback}, we reformulate OVD pseudo-labeling into multiple interpretable steps for robust pseudo-label generation.

\section{MSPL: Multi-Step Pseudo-Labeling}
\label{sec:method}
We introduce MSPL, an offline-to-online framework for robust pseudo-labeling in OVD, tailored for challenging scenarios such as crowding and occlusion. Unlike conventional single-step VLM alignment methods that rely on coarse caption-driven vocabulary and struggle with complex visual entanglement, MSPL performs structured multi-step reasoning to explicitly disentangle scene components. In the offline phase (\cref{sec:offline}), this reasoning is decomposed into three interpretable steps—object verification, label assignment, and background grounding—yielding robust pseudo-labels together with semantically rich intermediate representations. By decoupling structured reasoning from online, these representations serve as denoised supervision for online OVD training while substantially reducing online computational overhead. Building upon this supervision, the online phase (\cref{sec:online}) optimizes a contrastive objective that promotes generalization beyond base classes to potential unseen objects, grounds image-level captions at the region level, and alleviates background collapse in complex scenes.

\subsection{Three-Step Pseudo-Label (PL) Generation}
\label{sec:offline}

This section presents a three-step visual prompting framework for offline pseudo-labeling in OVD. To circumvent the aforementioned limitations of single-step VLM alignment, our approach leverages the synergy between SAM's foundational segmentation~\cite{SAM} and MLLM zero-shot reasoning. However, such direct integration is inherently prone to instability. Since SAM generates masks across diverse semantic granularities (\eg, whole objects, parts, or sub-parts), partial or non-object regions often receive erroneous semantic labels during MLLM reasoning. Furthermore, even for object-level regions, single-pass MLLM inference struggles in complex scenes where attention diffuses across overlapping objects or contextual distractors. Consequently, like single-step VLM alignment, this naive coupling inherits both proposal-level ambiguity and reasoning-level entanglement, often yielding inconsistent or hallucinated pseudo-labels.

To address these dual challenges, we impose structural constraints at both the region and reasoning levels. First, we restrict SAM outputs to whole-instance masks via hierarchical grouping~\cite{langsplat}, ensuring that subsequent reasoning operates on semantically coherent, object-level entities. Second, we regulate the interaction between each target region and its surrounding context by preserving the global scene structure while selectively attenuating non-target areas through controlled desaturation and blurring. As demonstrated by the visualizations of various modulations in the supplementary material, this visual context modulation stabilizes the visual evidence available for MLLM reasoning, mitigating distractions while retaining essential environmental cues. Although these measures enhance robustness relative to single-step VLM alignment, single-pass MLLM inference may still fall short in scenes with complex, overlapping, or heavily occluded objects. To address these residual ambiguities, we introduce a three-step reasoning process that sequentially verifies region validity, performs fine-grained category discovery, and disambiguates background elements.

\begin{figure}[tb]
  \centering
  \includegraphics[width=\linewidth]{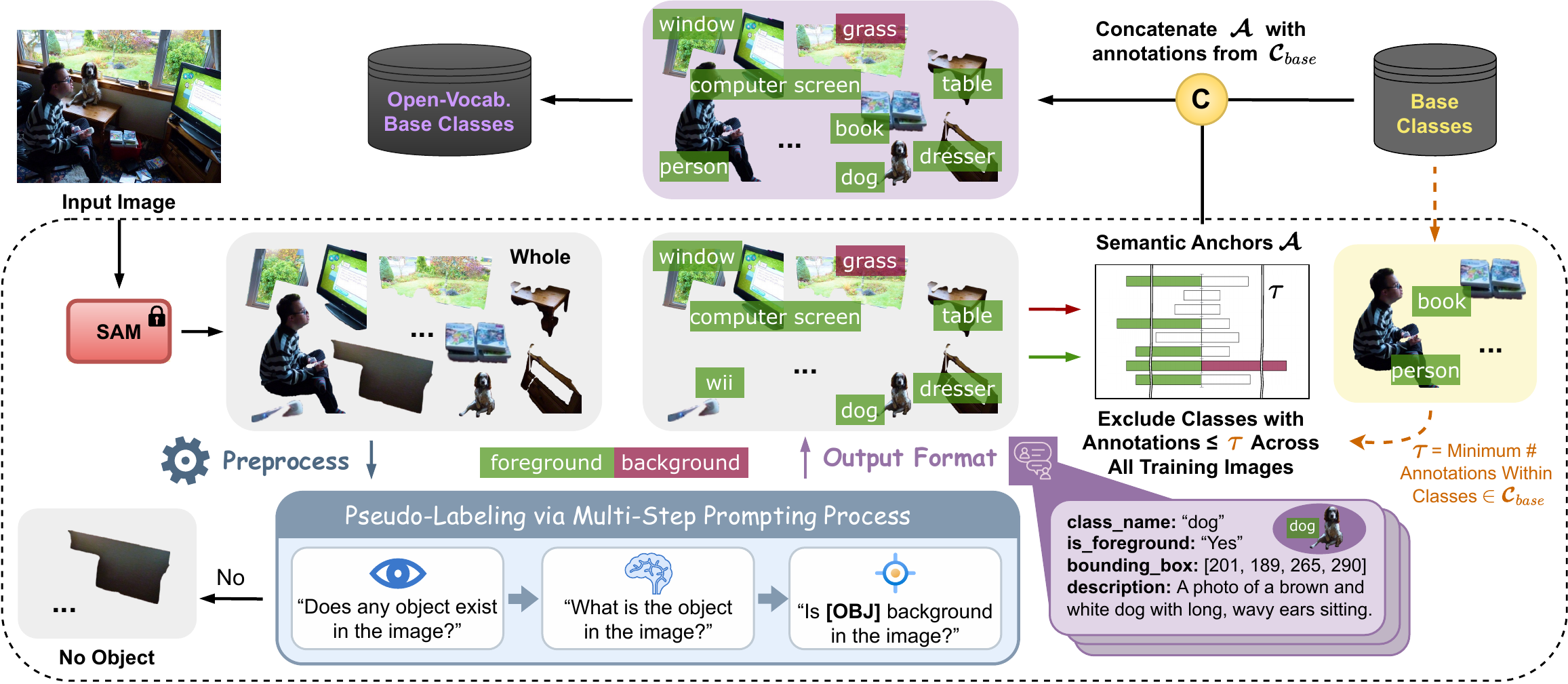}
  \caption{\textbf{Our offline multi-step pseudo-labeling.} We leverage foundational segmentation and MLLM-based zero-shot reasoning to implement a three-step framework (localization, recognition, grounding) with explicit intermediate reasoning goals. The resulting pseudo-labels are semantically refined and consolidated into the base dataset.}
 \label{fig:pl_generation}
\end{figure}

\subsubsection{Step 1: Pseudo-Box Verification}

Given the object-level region proposals, the first PL step verifies whether each region contains a valid object. Because SAM generates class-agnostic proposals, some regions may correspond to partial structures or non-object areas. Confirming object existence prior to semantic assignment restricts subsequent reasoning to visually grounded candidates, ensuring that later stages operate on reliable object regions. To this end, a robust MLLM~\cite{qwen} evaluates each region using the query ``\texttt{Does any object exist in the image?}'' and returns ``\texttt{Yes}'', ``\texttt{No}'', or ``\texttt{Unsure}'', forming a ternary decision filter. Regions confirmed as ``\texttt{Yes}'' proceed to the next reasoning step, whereas those labeled ``\texttt{No}'' or ``\texttt{Unsure}'' are discarded or recorded for explainability. As illustrated in~\cref{fig:pl_generation}, regions lacking discernible objects (\eg, plain dark areas) are removed. By narrowing the candidate set to visually valid objects, this step establishes a stable foundation for subsequent pseudo-label assignment.

\subsubsection{Step 2: Pseudo-Label Assignment.} 

Building upon the verified object regions from the previous step, we depart from conventional OVD paradigms that rely on single-step CLIP alignment against predefined class lexicons. Such approaches typically depend on vocabularies distilled from coarse image captions~\cite{cococaption}, which may omit or underspecify object details. To eliminate this vocabulary dependency, the second PL stage replaces rigid vocabulary-constrained alignment with zero-shot MLLM reasoning. By querying the model with category recognition, we explicitly elicit a category name along with a textual description for each validated region. In~\cref{fig:pl_generation}, this caption-agnostic formulation produces flexible pseudo-labels (\eg, ``dog'') while simultaneously generating region-level descriptions (\eg, ``A photo of a brown and white dog with long, wavy ears sitting.''). Importantly, any potential semantic ambiguities among the predicted labels are naturally resolved within the CLIP embedding space, where semantically related concepts occupy closely aligned representations, as detailed in~\cref{sec:online}. Such ambiguities include synonym distinctions, such as ``table'' vs. ``dining table'', and superclass variations, such as ``bird'' vs. ``parrot''. By leveraging the strong zero-shot recognition capabilities of MLLMs~\cite{gvt}, this step enables fine-grained object discovery beyond the static caption vocabulary.

\subsubsection{Step 3: Background Grounding.}

Despite the structural reasoning in the previous steps, residual ambiguities may persist, particularly when severely occluded objects yield ``\texttt{Unsure}’’ responses in the first PL step. Such objects remain unlabeled and are consequently assimilated into the learnable background class embedding during training~\cite{zsd,lbp}. In addition, background regions (\eg, ``tree'' or ``sky'') can be inadvertently labeled, generating noisy pseudo-labels that deviate from the objective of detecting foreground objects. This mislabeling biases the model toward background regions, which can interfere with learning discriminative features for semantically similar object classes. To address these issues, the third PL stage explicitly separates each category prediction into foreground and background decisions. Specifically, the model performs a binary verification to determine whether it corresponds to a true background concept (``\texttt{Yes}'' or ``\texttt{No}''). For example, in~\cref{fig:pl_generation}, the model identifies ``grass'' as background while retaining ``drawer'' as foreground. While denoising pseudo-labels, these reasoning decisions function as intermediate supervision that facilitates feature disentanglement during training, thereby enabling the recovery of object features that might otherwise be absorbed into background embeddings.

\subsubsection{Pseudo-label Refinement.}
\label{sec:method:postprocess}

Although the proposed three-step PL reasoning improves overall pseudo-label quality, spurious or hallucinated predictions may still occur due to inherent MLLM limitations. To enhance reliability, we introduce an MLLM-agnostic refinement strategy that filters pseudo-labels based on per-class prediction frequency. As demonstrated in~\cref{tab:ablation_semantic}, consistently predicted categories—such as clearly recognizable objects—accumulate high-frequency assignments, whereas ambiguous regions yield scattered predictions across classes, resulting in low frequencies. This observation motivates the use of per-class prediction frequency as a proxy for pseudo-label reliability. As illustrated in~\cref{fig:pl_generation}, pseudo-labels falling below a minimum threshold, derived from the reliable base-class distribution, are discarded. This ensures that the retained labels, termed \textit{semantic anchors}, exhibit support comparable to trusted categories. Finally, these anchors are integrated with the base classes to construct an open-vocabulary base set, providing reliable supervision for subsequent OVD training.

\begin{figure}[tb]
  \centering
  \includegraphics[width=\linewidth]{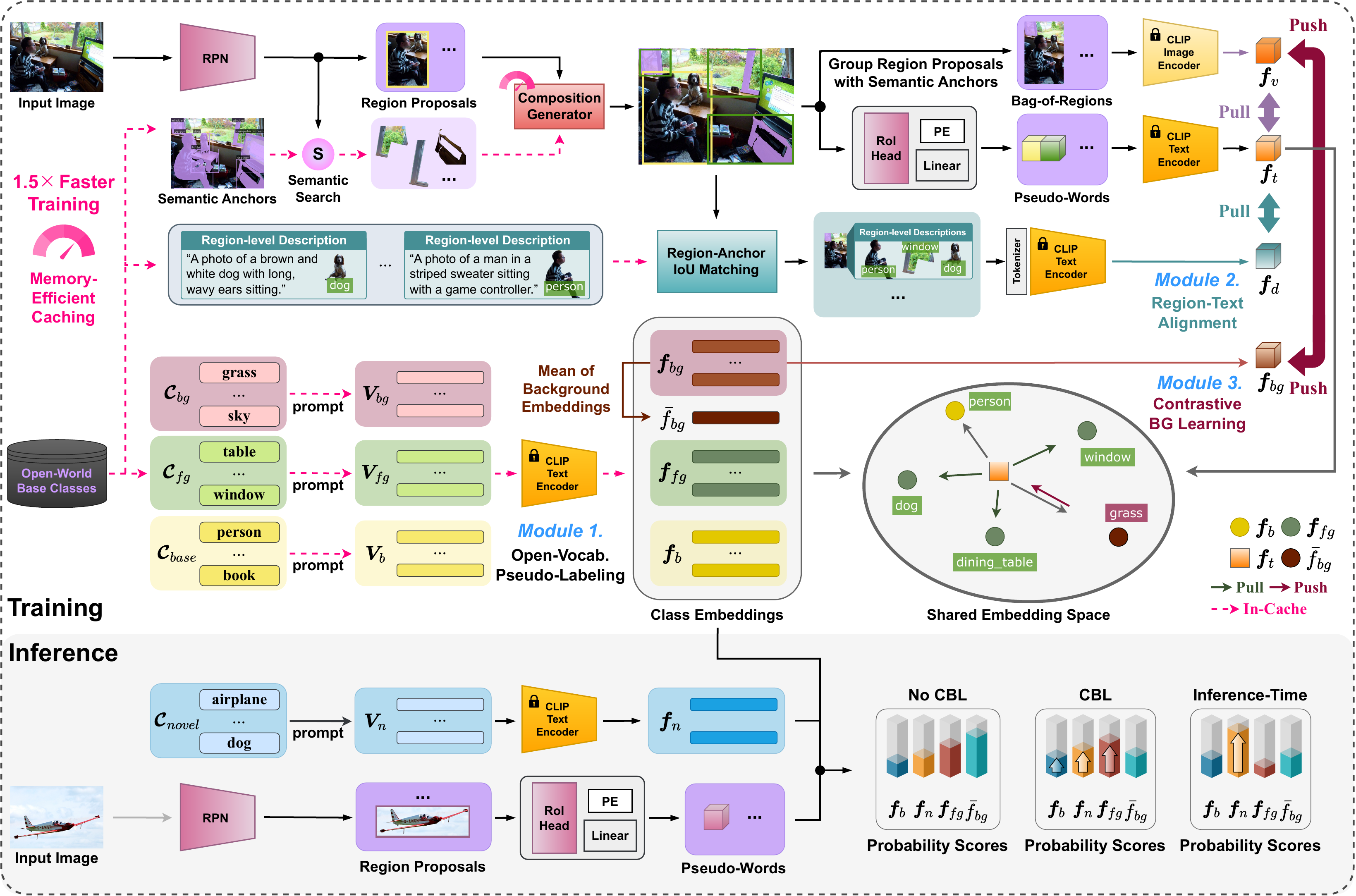}
  \caption{\textbf{Our online MSPL framework.} Module 1 performs pseudo-label-driven OVD; (2) Module 2 leverages description annotations; and (3) Module 3 utilizes background annotations to promote feature disentanglement. Memory caching accelerates composition augmentation, with pseudo-labels employed exclusively during training.}
  \label{fig:model}
\end{figure}

\subsection{Contrastive Learning with Multi-Step Supervision}
\label{sec:online}

This section details the online framework of MSPL, leveraging pre-computed offline supervision via a contrastive objective. As illustrated in~\cref{fig:model}, our architecture builds upon BARON~\cite{baron}, an effective OVD approach that captures rich contextual signals through the sampling of co-occurring objects. However, despite its strong performance, BARON suffers from computationally expensive online sampling, which creates a significant training bottleneck. To overcome this, we introduce an efficient composition generator that replaces online sampling with a \textit{bag-of-regions} grouping of cached semantic anchors, thereby accelerating training by a factor of 1.5. Within each bag, sampled region features are mapped into a joint word embedding space via a linear projection layer to form \textit{pseudo-word} embeddings. The text encoder $\mathcal{T}$ then processes these pseudo-words to generate a bag-of-regions text embedding $f_t^i = \mathcal{T}(w_0^i+ p_0^i, w_1^i+ p_1^i, \dots, w_{N^i-1}^i + p_{N^i-1}^i)$, where $N^i$ is the number of regions in the $i$-th bag, and $p_j^i$ is the learnable positional embedding for the $j$-th region. Finally, this text embedding is aligned with the corresponding visual embedding $f_v^i = \mathcal{V}(b_0^i, b_1^i, \dots, b_{N^i-1}^i)$ derived from the image encoder $\mathcal{V}$, with $b_j^i$ representing the visual feature of the $j$-th region.

For standard OV classification, each predicted region feature is assigned to the category yielding the highest CLIP cosine similarity. The candidate category set comprises both the original base classes $\boldsymbol{\mathcal{C}}_{base}$ and the foreground pseudo-labels $\boldsymbol{\mathcal{C}}_{fg}$, with their text embeddings pre-computed via the CLIP text encoder using prompt templates~\cite{vild,detic}. Given the continuous nature of the CLIP embedding space, aligning a region with a specific pseudo-label during training projects its feature into a shared semantic neighborhood. This naturally resolves synonym or superclass ambiguities (\eg, the novel class ``couch'' aligns closely with ``sofa''). Leveraging this property, our method employs pseudo-labels exclusively during training and discards them at inference to prevent misclassification.

\subsubsection{Region-Text Alignment (RTA).}
\label{sec:method:rta}
While basic visual features align with simple class names, such labels lack the descriptive richness required for fine-grained distinctions in complex scenes. For instance, detailed text can easily differentiate visually similar objects (\eg, a small red ``apple'' and a red ``ball''). To leverage this linguistic granularity, we propose RTA, which aligns pseudo-words with their corresponding region descriptions. Concretely, we define a matched set $\mathcal{S}$ of proposals with an IoU $> 0.7$ against pseudo-boxes. For each proposal in $\mathcal{S}$, its pseudo-word embedding $f_t^k$ and description embedding $f_d^k$ are aligned via the following objective~\cite{infonce}, with cosine similarity $\langle \cdot , \cdot \rangle$ and temperature $\tau$:
\begin{align}
    \mathcal{L}_{\text{RTA}} = - \frac{1}{|\mathcal{S}|} \sum_{k \in \mathcal{S}} \log \frac{\exp(\tau \cdot \langle f_t^k, f_d^k \rangle)}
    {\sum_{l \in \mathcal{S}} \exp(\tau \cdot \langle f_t^k, f_d^l \rangle)}.
    \label{eq:region_text}
\end{align}

\subsubsection{Contrastive Background Learning (CBL).}
\label{sec:method:cbl}
To mitigate background collapse, we propose a CBL strategy that explicitly disentangles all objects, including unlabeled ones, from background representations (\eg, ``sky'') in the feature space. In~\cref{fig:model}, the $B$ background concepts $\boldsymbol{\mathcal{C}}_{bg}$ identified during the third PL phase are encoded using the CLIP text encoder. These embeddings are treated as negative samples and averaged to initialize a learnable background prior $\bar{f}_{bg}$, which serves as a convergence target for true background features~\cite{infonce}:
\begin{align}
    \mathcal{L}_{\text{CBL}} = \frac{1}{2} \sum_{k=0}^{G-1} \left( \log p_{t,v}^k + \log p_{v,t}^k \right),
    \label{eq:loss}
\end{align}
where $G$ is the number of bags, and the $p_{t,v}^k$ and $p_{v,t}^k$ can be calculated as:
\begin{align}
    p_{t,v}^k &= \frac{\exp(\tau' \cdot \langle f_t^k, f_v^k \rangle)}
    {\sum_{l=0}^{G-1} \exp(\tau' \cdot \langle f_t^k, f_v^l \rangle) + \sum_{j=0}^{B-1} \exp(\tau'' \cdot \langle f_t^k, f_{bg}^j \rangle)}, \label{eq:img2text} \\
    p_{v,t}^k &= \frac{\exp(\tau' \cdot \langle f_v^k, f_t^k \rangle)}
    {\sum_{l=0}^{G-1} \exp(\tau' \cdot \langle f_v^k, f_t^l \rangle) + \sum_{j=0}^{B-1} \exp(\tau'' \cdot \langle f_v^k, f_{bg}^j \rangle)}, \label{eq:text2img}
\end{align}
where $\tau'$ and $\tau''$ are scaling factors. The loss promotes alignment of matched pairs and separation of background features, aiding foreground feature recovery.

\section{Experiments}
\label{sec:experiment}

\paragraph{Datasets and evaluation metrics.}
We evaluate MSPL on two widely used OVD benchmarks: OV-COCO~\cite{coco} and OV-LVIS~\cite{lvis}. Note that we generate pseudo annotations exclusively from base-category training images. For OV-COCO, we adopt the category split from OVR-CNN~\cite{ovrcnn}, dividing categories into 48 base and 17 novel classes. For OV-LVIS, following ViLD~\cite{vild}, we treat the 337 rare categories as novel and the common/frequent categories as base. Evaluation follows the OVR-CNN protocol: we report box AP at IoU 0.5 (AP$^{\text{N}}_{50}$) for novel categories on OV-COCO, and mask mAP (AP$_r$) for rare categories on OV-LVIS.
 
\paragraph{Implementation details.}
MSPL is implemented on Faster R-CNN~\cite{fasterrcnn} with a ResNet50-FPN backbone. Following recent works~\cite{baron, detpro}, the backbone is initialized with SOCO~\cite{soco} pre-trained weights and fine-tuned using synchronized batch normalization~\cite{syncbn} for a fair comparison. We employ $1\times$ and $2\times$ training schedules for OV-COCO and OV-LVIS, respectively. For mask generation in the offline phase, we adopt the SAM-B (ViT-B) model with default settings. As for the VLM, we utilize ViT-B-16~\cite{vit} with hand-crafted prompts from ViLD~\cite{vild} by default, adopting learned prompts~\cite{detpro} only for comparisons on OV-LVIS. The temperature scaling factors $\tau$, $\tau'$, and $\tau''$ are fixed at 0.2, 0.05, and 0.1. Following standard benchmarks~\cite{coco,occlusion}, we define \textit{Crowded} images as those with over eight objects and \textit{Occluded} instances as having over 50\% ground-truth box overlap. All other hyperparameters follow the settings of our baseline~\cite{baron}.

\begin{table}[!t]
    \centering
    \caption{\textbf{Results on OV-COCO}~\cite{coco}. Methods are grouped by additional supervision (\eg, weak or pseudo labels) beyond $\mathcal{C}_B$ instance labels. $\mathcal{C}_N$ denotes novel classes.}
    \label{tab:ovcoco}
    
    \begin{minipage}[t]{0.397\textwidth}
        \resizebox{\linewidth}{!}{%
            \begin{tabular}[t]{l l c >{\color{tablegray}}c}
                \toprule
                \textbf{Methods} & \textbf{Backbone} & $\text{AP}_{50}^{N}$  & $\text{AP}_{50}^{B}$  \\
                \midrule
                \multicolumn{4}{c}{Instance labels in $\mathcal{C}_B$ (CLIP Supervision)} \\
                \midrule
                ViLD-ens~\cite{vild}                     & RN50 (24M)        & 27.6  & 51.3 \\
                BARON~\cite{baron}                       & RN50 (24M)        & 34.0  & 60.4 \\
                CORA~\cite{cora}                         & RN50 (24M)        & 35.1  & 35.4 \\
                BIND~\cite{bind}                         & ViT-B/16 (86M)    & 36.3  & 50.2 \\
                CLIP-Self~\cite{clipself}                & ViT-B/16 (86M)    & 37.6  & - \\
                LBP~\cite{lbp}                           & RN50 (24M)        & 37.8  & 58.7 \\
                CCKT-Det~\cite{ccktdet}                  & RN50 (24M)        & 38.0  & 35.0 \\
                CAKE~\cite{cake}                         & RN50 (24M)        & 38.2  & - \\
                OV-DQUO~\cite{ovdquo}                    & RN50 (24M)        & 39.2  & - \\ 
                DeCo-DETR~\cite{decodetr}                & RN50 (24M)        & 41.3  & - \\ \hdashline
                BIND~\cite{bind}                         & ViT-L/16 (307M)   & 41.5  & 54.8 \\
                CCKT-Det~\cite{ccktdet}                  & SwinB (88M)       & 41.9  & 40.9 \\
                CORA+~\cite{cora}                        & RN50$\times$4 (87M) & 43.4  & 43.8 \\
                CLIP-Self~\cite{clipself}                & ViT-L/14 (307M)   & 44.3  & - \\
                OV-DQUO~\cite{ovdquo}                    & RN50$\times$4 (87M) & 45.6  & - \\
                \bottomrule
            \end{tabular}%
        }
    \end{minipage}\hfill
    \begin{minipage}[t]{0.583\textwidth}
        \resizebox{\linewidth}{!}{%
            \begin{tabular}[t]{l l l c >{\color{tablegray}}c}
                \toprule
                \textbf{Methods} & \textbf{Supervision} & \textbf{Backbone} & $\text{AP}_{50}^{N}$  & $\text{AP}_{50}^{B}$  \\
                \midrule
                \multicolumn{5}{c}{Extra caption datasets, Weak/Pseudo Labels in $\mathcal{C}_B \cup \mathcal{C}_N$} \\
                \midrule
                Detic~\cite{detic}                       & IN21K \& CC3M     & RN50 (24M)        & 27.8  & 42.0 \\
                OV-DETR~\cite{ovdetr}                    & Pseudo annotations      & RN50 (24M)        & 29.4  & 52.7 \\
                CoDet~\cite{codet}                       & CC3M \& COCO Caption  & RN50 (24M)        & 30.6  & 46.4 \\
                PB-OVD~\cite{PB-OVD}                     & COCO Caption          & RN50 (24M)        & 30.8  & 46.4 \\
                VL-PLM~\cite{VL-PLM}                     & Pseudo annotations      & RN50 (24M)        & 34.4  & 60.2 \\
                RegionCLIP~\cite{RegionCLIP}             & CC3M              & RN50 (24M)        & 35.2  & 57.6 \\
                OC-OVD~\cite{ocovd}                      & COCO Caption          & RN50 (24M)        & 36.6  & 49.4 \\
                SAS-Det~\cite{sasdet}                    & COCO Caption          & RN50 (24M)        & 37.4  & 58.5 \\
                DITO~\cite{dito}                         & LAION-2B          & ViT-B/16(86M)     & 36.6  & 48.8 \\
                LP-OVOD~\cite{LP-OVOD}                         & Pseudo annotations       & RN50 (24M)    & 40.5  & 60.5 \\
                \rowcolor{gold} 
                \textbf{MSPL (Ours)}                   & Pseudo annotations      & RN50 (24M)        & \textbf{43.4} & 58.9 \\ \hdashline
                CFM-ViT~\cite{cfmvit}                    & LAION-2B          & ViT-L/16 (307M)   & 34.3  & 46.4 \\
                RegionCLIP~\cite{RegionCLIP}             & CC3M              & RN50$\times$4 (87M) & 39.3  & 61.6 \\
                DITO~\cite{dito}                         & DataComp-1B       & ViT-L/16(307M)    & 40.2  & 54.6 \\
                CORA+~\cite{cora}                         & COCO Caption       & RN50$\times$4 (87M)    & 43.1  & 56.2 \\
                \rowcolor{gold} 
                \textbf{MSPL (Ours)}                   & Pseudo annotations      & RN50$\times$4 (87M) & \textbf{47.8} & 60.9 \\
                \bottomrule
            \end{tabular}%
        }
    \end{minipage}
    
\end{table}
\begin{table}[!t]
    \centering
    
    \begin{minipage}[t]{0.4625\textwidth}
        \caption{Statistics of pseudo-labels.}
        \label{tab:stat}
        
        \renewcommand{\arraystretch}{1.35} 
        
        \resizebox{\linewidth}{!}{%
            \begin{tabular}{l r r r}
                \toprule
                \textbf{Metric} & \textbf{BLIP2} & \textbf{InstructBLIP} & \textbf{Qwen2} \\
                \midrule
                \multicolumn{4}{l}{\textbf{Total}} \\
                \quad \# Classes & 6.0K & 3.1K & 3.9K \\
                \quad \# Annotations & 395K & 567K & 637K \\
                \quad \# ``\texttt{Unsure}'' & 1.5M & 1.1M & 563K \\
                \midrule
                \multicolumn{4}{l}{\textbf{OV-COCO} (17 novel classes only)} \\
                \quad \# Classes & 31 & 30 & 65 \\
                \quad \# Annotations & 197K & 294K & 202K \\
                \rowcolor{cyan!25} 
                Hard Hit (\%) & 41.2 & \textbf{47.1} & \textbf{47.1} \\
                \rowcolor{cyan!10} 
                Soft Hit (\%) & 85.0 & 81.8 & \textbf{86.0} \\
                \midrule
                \multicolumn{4}{l}{\textbf{OV-LVIS} (337 rare classes only)} \\
                \quad \# Classes & 5.3K & 2.5K & 3.3K \\
                \quad \# Annotations & 137K & 315K & 232K \\
                \rowcolor{cyan!25} 
                Hard Hit (\%) & 31.1 & 27.6 & \textbf{34.1} \\
                \rowcolor{cyan!10} 
                Soft Hit (\%) & 77.9 & \textbf{85.7} & 85.3 \\
                \bottomrule
            \end{tabular}%
        }
    \end{minipage}\hfill
    \begin{minipage}[t]{0.5175\textwidth}
        \caption{Comparison on OV-LVIS.}
        \label{tab:ov_lvis}
        
        \renewcommand{\arraystretch}{1.15} 
        
        \resizebox{\linewidth}{!}{%
            \begin{tabular}{l c>{\color{tablegray}}c>{\color{tablegray}}c>{\color{tablegray}}c c>{\color{tablegray}}c>{\color{tablegray}}c>{\color{tablegray}}c}
                \toprule
                \multirow{2}{*}{\textbf{Method}} & \multicolumn{4}{c}{\textbf{Detection}} & \multicolumn{4}{c}{\textbf{Segmentation}} \\
                \cmidrule(lr){2-5} \cmidrule(lr){6-9}
                 & $AP_r$ & $AP_c$ & $AP_f$ & AP & $AP_r$ & $AP_c$ & $AP_f$ & AP \\
                \midrule
                ViLD~\cite{vild} & 16.7 & 26.5 & 34.2 & 27.8 & 16.6 & 24.6 & 30.3 & 25.5 \\
                RegionCLIP~\cite{RegionCLIP} & 17.1 & 27.4 & 34.0 & 28.2 & - & - & - & - \\
                CCKT-Det++~\cite{ccktdet} & 18.2 & - & - & 27.1 & - & - & - & - \\
                OV-DETR~\cite{ovdetr} & - & - & - & - & 17.4 & 25.0 & 32.5 & 26.6 \\
                VLDet~\cite{vldet} & - & - & - & - & 21.7 & 29.8 & 34.3 & 30.1 \\
                Detic~\cite{detic} & - & - & - & - & 17.8 & 26.3 & 31.6 & 26.8 \\
                MIC~\cite{mic} & 22.9 & 34.0 & 39.9 & 34.4 & 20.8 & \textbf{30.5} & \textbf{35.4} & \textbf{30.7} \\
                DetPro~\cite{detpro} & 20.8 & 27.8 & 32.4 & 28.4 & 19.8 & 25.6 & 28.9 & 25.9 \\
                OC-OVD~\cite{ocovd} & 21.1 & 25.0 & 29.1 & 25.9 & - & - & - & - \\
                OADP~\cite{oadp} & 21.9 & 28.4 & 32.0 & 28.7 & 21.7 & 26.3 & 29.0 & 26.6 \\
                DK-DETR~\cite{dkdetr} & 22.2 & 32.0 & \textbf{40.2} & 33.5 & 20.5 & 28.9 & \textbf{35.4} & 30.0 \\
                BARON~\cite{baron} & 23.2 & 29.3 & 32.5 & 29.5 & 22.6 & 27.6 & 29.8 & 27.6 \\
                CoDet~\cite{codet} & 23.4 & 30.0 & 34.6 & 30.7 & - & - & - & - \\
                LBP~\cite{lbp} & 24.1 & 29.5 & 32.8 & 29.9 & 23.7 & 27.7 & 30.1 & 28.0 \\
                CAKE~\cite{cake} & 25.0 & \textbf{34.8} & 38.4 & \textbf{34.9} & 23.9 & 29.1 & 33.6 & 28.7 \\
                BIRDet~\cite{birdet} & 26.0 & 21.7 & 29.5 & 25.5 & - & - & - & - \\
                RALF~\cite{ralf} & 21.9 & 26.2 & 29.1 & 26.6 & - & - & - & - \\
                \rowcolor{gold} 
                \textbf{MSPL (Ours)} & \textbf{26.4} & \textbf{34.8} & 38.2 & \textbf{34.9} & \textbf{24.8} & 28.5 & 33.0 & 28.6 \\
                \bottomrule
            \end{tabular}%
        }
    \end{minipage}
    
\end{table}

\subsection{Main Results}
\subsubsection{Comparison with state-of-the-art methods.}
We compare MSPL against state-of-the-art OVD methods on the OV-COCO and OV-LVIS benchmarks. In~\cref{tab:ovcoco}, MSPL establishes a new state-of-the-art on OV-COCO among recent methods leveraging auxiliary datasets for pseudo-annotations~\cite{codet,RegionCLIP,ovdetr,LP-OVOD}. Specifically, it achieves 43.4 and 47.8 $\text{AP}_{50}^{N}$ for ResNet-50 and ResNet-50$\times$4 backbones, respectively. Notably, our approach consistently outperforms distillation-based methods relying strictly on CLIP knowledge and base-class labels, surpassing the recent leading DeCo-DETR~\cite{decodetr} by 2.1 $\text{AP}_{50}^{N}$. On the challenging OV-LVIS benchmark in~\cref{tab:ov_lvis}, MSPL sets a new record for rare categories with a detection $\text{AP}_r$ of 26.4 and a segmentation $\text{AP}_r$ of 24.8. For detection, MSPL outperforms BIRDet~\cite{birdet} by 0.4 $\text{AP}_r$ and surpasses CAKE~\cite{cake} by 1.4 $\text{AP}_r$. In instance segmentation, MSPL exceeds strong baselines~\cite{baron,lbp,cake} by significant margins. This consistent superiority across benchmarks confirms that our proposed strategies scale exceptionally well to large-vocabulary datasets.

\subsubsection{Statistics.}
\cref{tab:stat} compares pseudo-label statistics across MLLM variants. Notably, Qwen2~\cite{qwen} produces the densest and most confident annotations—637K across 3.9K categories—while yielding minimal ``Unsure'' responses. To evaluate coverage of unseen domains, we measure both \textit{Hard Hit} (exact string match, \eg, ``cup'' to ``cup'') and \textit{Soft Hit} (CLIP cosine similarity $> 0.8$ following~\cite{marvelovd}, \eg, ``vehicle'' to ``bus''). Although Hard Hit provides direct supervision for OVD, its maximum coverage is inherently limited by synonym and superclass ambiguities. Since open-vocabulary detectors operate within a continuous CLIP embedding space, exact textual matches are not strictly necessary. Pseudo-labels comprising synonyms or semantically related terms remain closely aligned with the corresponding novel classes, thereby providing effective supervision. Under this broader semantic criterion, Qwen2 achieves strong Soft Hit rates of 86.0\% and 85.3\% on OV-COCO and OV-LVIS, respectively. These results demonstrate the large scale, vocabulary diversity, and broad semantic coverage of our pseudo-labels, supporting their suitability as supervision for OVD.

\subsubsection{Pseudo-label analysis.}
\cref{tab:pl} evaluates the trade-off between pseudo-labeling efficiency and quality under varying scene complexities, specifically in \textit{Crowded} and \textit{Occluded} settings. Prior methods~\cite{PB-OVD,VL-PLM,sasdet} rely on single-step VLM alignment and are susceptible to visual interference, resulting in suboptimal pseudo-labels. On a single A6000 GPU, SAS-Det~\cite{sasdet} achieves a per-image generation time of 0.13s after training; however, its online self-training increases the total cost to over one second per image. In contrast, our method follows the offline paradigm~\cite{PB-OVD,VL-PLM}, performing intensive VLM or MLLM inference only once during pseudo-label generation while avoiding iterative and expensive online self-training. This design improves overall throughput for large-scale labeling. By combining a segmentation model~\cite{SAM} with asynchronous MLLM inference via batch parallelism, our method achieves a favorable accuracy–efficiency balance, averaging 0.43s per image on a single GPU and processing the COCO training set in approximately 5 hours using multi-threaded execution across 8 GPUs.

\begin{table}[tb]
    \centering
    
    \begin{minipage}[t]{0.39\textwidth}
        \caption{Component ablation.}
        \label{tab:ablation}
        
        \resizebox{\linewidth}{!}{%
            {\renewcommand{\arraystretch}{1.5}
            \begin{tabular}{c c c c c}
                \toprule
                \multicolumn{2}{c}{Multi-Step Pseudo-Label} & \multirow{1}{*}{RTA} & \multirow{1}{*}{CBL} & \multirow{1}{*}{$\text{AP}_{50}^{N}$} \\
                \cmidrule(lr){1-2}
                \multicolumn{1}{c}{1-Step} & \multicolumn{1}{c}{3-Step} & & & \\
                \midrule
                - & - & - & - & 34.0 \\
                \midrule
                \checkmark & - & - & - & 37.6 \\
                - & \checkmark & - & - & 41.6 \\ \hdashline
                - & \checkmark & \checkmark & - & 42.5 \\
                \rowcolor{gold} 
                - & \checkmark & \checkmark & \checkmark & \textbf{43.4} \\
                \bottomrule
            \end{tabular}%
            }
        }
    \end{minipage}\hfill
    \begin{minipage}[t]{0.59\textwidth}
        \caption{Per-image pseudo-labeling time and novel-class quality ($\text{AP}_{50}^{N}$) on the OV-COCO validation set using a single A6000 GPU.}
        \label{tab:pl}
        
        \resizebox{\linewidth}{!}{%
            {\renewcommand{\arraystretch}{1.51}
            \begin{tabular}{l cccc}
                \toprule
                Method & OV-COCO & \textit{Crowded} & \textit{Occluded} & Time (s) \\
                \midrule
                PB-OVD~\cite{PB-OVD}   & 18.7 & 5.1 & 2.7 & 0.49 \\
                VL-PLM~\cite{VL-PLM}   & 25.5 & 7.3 & 3.8 & 0.45 \\
                SAS-Det~\cite{sasdet}  & 26.7 & 11.6 & 5.7 & $\gg1.0$ \\
                \rowcolor{gold} 
                \textbf{MSPL (Ours)}   & \textbf{32.3} & \textbf{23.9} & \textbf{15.5} & \textbf{0.43} \\
                \bottomrule
            \end{tabular}%
            }
        }
    \end{minipage}
\end{table}

\subsection{Ablation Analysis}

\subsubsection{Impact of individual modules.}
\cref{tab:ablation} evaluates the incremental contribution of each component in the MSPL framework. The one-step variant directly predicts pseudo-labels without intermediate reasoning supervision, yet it still improves over the baseline. This gain arises from our structured integration of SAM and MLLM---specifically, object-level SAM masks and visual context modulation---which alleviates the limitations of naive model coupling. Building on this foundation, transitioning to three-step PL reasoning yields larger gains, indicating that sequential decomposition with intermediate supervision is more robust for parsing complex scenes than single-pass prompting. RTA further refines label quality by incorporating fine-grained linguistic attributes, which are essential for disambiguating semantically similar categories. Finally, CBL effectively mitigates background collapse, resulting in additional performance gains. Collectively, these results highlight the complementary roles of structured reasoning and semantically rich supervision in achieving high-fidelity pseudo-labeling.

\begin{table*}[t]
    \centering
    
    \begin{minipage}[t]{0.74\textwidth}
        \caption{Zero-shot transfer on MS-COCO and Objects365~\cite{object365} with models trained on OV-LVIS without fine-tuning.}
        \label{tab:transfer}
        \centering
        \resizebox{\linewidth}{!}{%
            \begin{tabular}{l | ccc >{\color{tablegray}}c >{\color{tablegray}}c >{\color{tablegray}}c | ccc >{\color{tablegray}}c >{\color{tablegray}}c >{\color{tablegray}}c}
                \toprule
                & \multicolumn{6}{c|}{MS-COCO~\cite{coco}} & \multicolumn{6}{c}{Objects365~\cite{object365}} \\
                \textbf{Methods} & $AP$ & $AP_{50}$ & $AP_{75}$ & $AP_s$ & $AP_m$ & $AP_l$ & $AP$ & $AP_{50}$ & $AP_{75}$ & $AP_s$ & $AP_m$ & $AP_l$ \\ \midrule
                Supervised & 46.5 & 67.6 & 50.9 & 27.1 & 67.6 & 77.7 & 25.6 & 38.6 & 28.0 & 16.0 & 28.1 & 36.7 \\ \midrule
                ViLD~\cite{vild} & 34.1 & 52.3 & 36.5 & 21.6 & 38.9 & 46.1 & 11.5 & 17.8 & 12.3 & 4.2 & 11.1 & 17.8 \\
                DetPro~\cite{detpro} & 34.9 & 53.8 & 37.4 & 22.5 & 39.6 & 46.3 & 12.1 & 18.8 & 12.9 & 4.5 & 11.5 & 18.6 \\
                BARON~\cite{baron} & 36.2 & 55.7 & 39.1 & 24.8 & 40.2 & 47.3 & 13.6 & 21.0 & 14.5 & 5.0 & 13.1 & 20.7 \\
                LBP~\cite{lbp} & 36.8 & 56.5 & 39.8 & 25.6 & 40.6 & 48.1 & 14.3 & 21.8 & 15.1 & 5.5 & 13.7 & 21.6 \\ 
                \rowcolor{gold}
                \textbf{MSPL (Ours)} & \textbf{37.5} & \textbf{57.4} & \textbf{40.8} & \textbf{26.4} & \textbf{41.4} & \textbf{49.1} & \textbf{15.1} & \textbf{22.7} & \textbf{15.9} & \textbf{6.1} & \textbf{14.4} & \textbf{22.5} \\ \bottomrule
            \end{tabular}%
        }
    \end{minipage}\hfill
    \begin{minipage}[t]{0.24\textwidth}
        \caption{Thresholds of semantic anchors.}
        \label{tab:ablation_semantic}
        \centering
        \resizebox{0.8\linewidth}{!}{%
            {\renewcommand{\arraystretch}{1.14}
            \begin{tabular}{c c}
                \toprule
                Threshold & $\text{AP}^{N}_{50}$ \\
                \midrule
                \texttt{RANDOM} & 41.7 \\
                \texttt{ALL} & 42.1 \\
                \texttt{MEDIUM} & 42.5 \\
                \texttt{AVERAGE} & 42.6 \\
                \texttt{MIN} & \textbf{43.4} \\
                \bottomrule
            \end{tabular}%
            }
        }
    \end{minipage}
\end{table*}
\begin{table}[tb]
    \centering
    
    \begin{minipage}[t]{0.36\textwidth}
        \caption{Impact of generators.}
        \label{tab:pg}
        \centering
        \resizebox{0.75\linewidth}{!}{%
            \begin{tabular}{l c}
                \toprule
                Proposal generator & $AP_{50}^{N}$ \\
                \midrule
                Mask R-CNN~\cite{maskrcnn} & 40.9 \\
                MAVL~\cite{mavl}           & 42.2  \\
                SAM~\cite{SAM}            & \textbf{43.4} \\
                \bottomrule
            \end{tabular}%
        }
    \end{minipage}\hfill
    \begin{minipage}[t]{0.338\textwidth}
        \caption{MLLM variants.}
        \label{tab:mllm}
        \centering
        \resizebox{0.89\linewidth}{!}{%
            \begin{tabular}{l c c}
                \toprule
                Model & Size & $AP_{50}^{N}$ \\
                \midrule
                BLIP2~\cite{blip2}        & 2.7B & 39.6 \\
                InstructBLIP~\cite{instructblip} & 7B  & 42.6 \\
                Qwen2~\cite{qwen}         & 7B  & \textbf{43.4} \\
                \bottomrule
            \end{tabular}%
        }
    \end{minipage}\hfill
    \begin{minipage}[t]{0.3\textwidth}
        \caption{Visual context.}
        \label{tab:preprocessing}
        \centering
        \resizebox{0.725\linewidth}{!}{%
            \begin{tabular}{l c}
                \toprule
                Strategy & $AP_{50}^{N}$ \\
                \midrule
                Bounding box    & 33.2 \\
                Black mask      & 38.7 \\
                Blur \& gray    & \textbf{43.4} \\
                \bottomrule
            \end{tabular}%
        }
    \end{minipage}
\end{table}

\subsubsection{Transfer to the other datasets.}
Following standard protocols~\cite{vild,baron}, we evaluate the zero-shot transfer capability of MSPL by evaluating a model trained on OV-LVIS across two benchmarks: MS-COCO~\cite{coco} and Objects365~\cite{object365}. In~\cref{tab:transfer}, MSPL achieves superior performance across all datasets, outperforming existing state-of-the-art methods. This robust generalization across diverse data distributions underscores the broad applicability of our approach in real-world scenarios.

\subsubsection{Impact of semantic anchors.}
\cref{tab:ablation_semantic} evaluates semantic anchor policies based on base-class statistics. The \texttt{RANDOM} (70\% sampling) and \texttt{ALL} policies yield the lowest performance, as they disregard per-class annotation frequency derived from the base-class statistics. The improved results of \texttt{AVERAGE} and \texttt{MEDIAN} indicate that lower annotation counts are associated with less reliable anchors. Accordingly, \texttt{MIN} achieves the best performance by using the minimum base-class frequency as a strict cutoff to filter noisy or hallucinated anchors.

\subsubsection{Impact of proposal generators.}
In~\cref{tab:pg}, we evaluate different class-agnostic proposal generators. We observe that SAM~\cite{SAM} provides higher-quality pseudo-box candidates by localizing arbitrary objects beyond closed-set vocabularies~\cite{maskrcnn,mavl}. The consistent gains across generators indicate that our multi-step reasoning is robust and benefits from stronger open-world localization.

\subsubsection{Impact of MLLM variants.}
\cref{tab:mllm} analyzes the impact of MLLM scale on pseudo-label quality. While performance remains comparable among models within the same parameter tier (\eg, 7B variants), we observe an approximately linear improvement as the scale increases from 2.7B to 7B. Notably, even the compact 2.7B model~\cite{blip2} yields substantial gains over the baseline, demonstrating effectiveness in resource-constrained settings. These results indicate that our framework is robust to architectural differences at a fixed scale, while reliably translating increased model capacity into improved label fidelity.

\subsubsection{Impact of context modulation.}
We examine preprocessing strategies to reduce MLLM sensitivity to visual context. Raw proposals often degrade reasoning accuracy, while masking non-target regions improves focus but can cause hallucination from context loss and silhouette artifacts. In contrast, our blurred-and-grayscale strategy suppresses background interference while preserving essential context, achieving the best performance. This suggests that balancing target emphasis and contextual preservation is crucial for high-fidelity pseudo-labeling.

\subsubsection{Further analysis.}
\begin{figure*}[t]
  \centering
  
  \begin{minipage}[t]{0.39\textwidth}
    \centering
    \includegraphics[width=\linewidth]{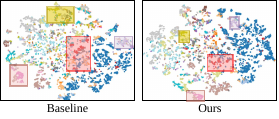}
    \caption{t-SNE results of features.}
    \label{fig:tsne}
  \end{minipage}
  \hfill
  \begin{minipage}[t]{0.57\textwidth}
    \centering
    \includegraphics[width=\linewidth]{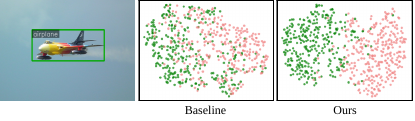}
    \caption{t-SNE results of background separation.}
    \label{fig:background}
  \end{minipage}
\end{figure*}

We use t-SNE~\cite{tsne} to visualize feature distributions and analyze background interpretation. As shown in~\cref{fig:tsne}, MSPL learns more compact and discriminative representations for novel categories than the baseline~\cite{baron}. We further examine background–foreground disentanglement for the ``airplane'' class in~\cref{fig:background}, where MSPL forms a distinct cluster well separated from learnable background embeddings, while the baseline shows substantial overlap.

\section{Conclusion}
\label{sec:conclusion}
In this paper, we introduce MSPL, a multi-step pseudo-labeling framework for open-vocabulary object detection (OVD) that reformulates complex scene understanding through multi-step visual reasoning. By decomposing labeling into three interpretable stages—object localization, category recognition, and background grounding—MSPL addresses the limitations of single-step vision–language alignment, particularly in crowded or occluded scenes. The resulting intermediate representations provide semantically enriched supervision for contrastive learning: region–text alignment incorporates fine-grained linguistic attributes, while contrastive background learning alleviates background collapse. Extensive evaluations on OVD benchmarks demonstrate that MSPL achieves state-of-the-art performance, scales effectively with multimodal large language model capacity, and maintains strong pseudo-label quality and efficiency.
We hope our work inspires further exploration of multi-step visual reasoning in perception tasks. 


%
%
\bibliographystyle{splncs04}
\bibliography{main}

\title{MSPL: Multi-Step Pseudo-Labeling for Open-Vocabulary Object Detection\\(Supplementary Material)} 

\titlerunning{MSPL}

\author{Hojun Choi\inst{1}\orcidlink{0009-0009-4121-4519} \and
Youngsun Lim\inst{2}\orcidlink{0009-0006-8791-0964} \and
Jaeyo Shin\inst{1}\orcidlink{0009-0005-3931-938X} \and
Hyunjung Shim\inst{1}\orcidlink{0000-0001-6796-1058}}

\authorrunning{H.~Choi et al.}

\institute{KAIST AI, South Korea\\
\email{\{hchoi256,jaeyo\_shin,kateshim\}@kaist.ac.kr} \and
Boston University, USA\\
\email{youngsun@bu.edu}}

\maketitle

\renewcommand{\thesection}{\Alph{section}}
\renewcommand{\thetable}{A\arabic{table}}
\renewcommand{\thefigure}{A\arabic{figure}}
\renewcommand{\theequation}{A\arabic{equation}}

\section*{Contents}
\noindent
A. Device Information \dotfill 22\\
B. Limitations \& Future Work \dotfill 22\\
C. Implementation Details \dotfill 24\\
D. Hyperparameters \dotfill 26\\
E. Additional Ablation Study \dotfill 26\\
\hspace*{1em}E.1 Visual Context Modulation \dotfill 26\\
\hspace*{1em}E.2 Pseudo-Label Statistics \dotfill 28\\
F. Additional Qualitative Results \dotfill 29\\
\hspace*{1em}F.1 Pseudo-Label Visualization \dotfill 29\\
\hspace*{1em}F.2 Detection Visualization \dotfill 30\\
G. Specifications of Baseline Models \dotfill 34\\
H. Prompt Templates \dotfill 36

\section{Device Information}
\label{appendix:device}
All experiments were implemented in PyTorch 1.12.1 and conducted on a workstation equipped with eight NVIDIA A6000 GPUs. The online training phase required approximately 11 hours with a peak VRAM utilization of 12,000 GB per GPU. In contrast, the offline stage---leveraging our efficient asynchronous design across the entire COCO benchmark---required roughly 5 hours to complete. This offline execution utilized the full GPU memory capacity via multi-threaded processing across all 8 GPUs. To facilitate a fair comparison and ensure reproducibility, we fixed the random seed to 1194806617 for all experimental trials.

\section{Limitations \& Future Work}
\label{appendix:limitations}
In this section, we analyze potential failure scenarios for our framework: (L1) inherent MLLM dependencies, (L2) extreme long-tail cases, and (L3) temporal (4D) integration. Exploring these issues serves as a promising direction for future work within the community.

\subsection{Scaling Dependency on MLLM Capabilities}
MSPL’s performance is intrinsically tied to the reasoning capabilities of the underlying MLLM; less capable models tend to generate lower-quality pseudo-labels, which can degrade open-vocabulary detection (OVD) performance. To mitigate this scale-dependency, MSPL employs a tri-modal judgment mechanism (``\texttt{Yes},'' ``\texttt{No},'' or ``\texttt{Unsure}'') as a hard gate, as detailed in~\cref{sec:offline}. Each generated response, accompanied by its corresponding reasoning rationale, is systematically analyzed and visualized in~\cref{ablation:qualitative}. Therefore, this conservative reasoning heuristic enables the MLLM to discard potentially false-positive predictions labeled as ``\texttt{Unsure}.'' Consequently, even small-scale MLLMs (\eg, 2.7B parameters) can produce higher-fidelity pseudo-labels than existing methods~\cite{PB-OVD,VL-PLM,sasdet}, as validated in~\cref{tab:mllm}. Nevertheless, while MSPL improves the performance baseline for smaller models, we recommend at least 7B-scale MLLMs to fully leverage the representative power of our multi-step PL framework.

\subsection{Challenges in Long-Tailed Scenarios}
While MSPL is robust, it does not yet fully utilize ``\texttt{Unsure}'' responses---a rich potential source of long-tailed labels---to positively contribute to OVD training. Since false-positive predictions are detrimental to overall performance, as shown in~\cref{tab:ablation,tab:mllm,tab:ablation_semantic}, we prioritize filtering these instances rather than directly incorporating them---a cornerstone of our safety-preserving, high-fidelity labeling strategy. To provide the community with actionable insights, we identify two primary dimensions where long-tailed challenges manifest within our framework.

\subsubsection{Color-agnostic visual ambiguity.}
The ``\texttt{Unsure}'' responses from a robust MLLM provide a direct signal regarding which instances are visually distracting or ambiguous. Our analysis reveals a primary culprit: MLLMs often over-rely on texture or shape at the expense of color. For instance, any region that shares a silhouette with a ``knife'' is often misclassified as such, even when its color profile is entirely incongruent---a known limitation in current vision-language models (VLMs)~\cite{colorbench}. MSPL remains susceptible to these inherent color-awareness deficits. In this regard, we provide an interpretable analysis of the MLLM’s evidence for each pseudo-label, with the expectation that performance will scale in tandem with future advancements in chromatic-aware MLLMs.

\subsubsection{Linguistic and semantic granularity.}
Long-tailed categories are often filtered out by frequency thresholds applied to base classes, as discussed in~\cref{tab:ablation_semantic}, where low-count annotations are discarded as noise. These abandoned labels often represent objects with high semantic granularity or proper nouns (\eg, ``Eiffel Tower'' or ``Mount Everest'') rather than generic class names (\eg, ``building'' or ``mountain''). Due to their sparse representation within the training set, these categories manifest as low-frequency outliers characterized by minimal per-class annotation counts. This suggests that MLLMs struggle with hierarchical linguistic granularities when collapsing complex visual reasoning into a single class prediction. Recent work~\cite{shine} highlights this by establishing hierarchical semantic structures within object categories. Rather than erroneously mapping a detection to a specific proper noun (\eg, ``Hogwarts Castle''), which limits generalizability, this approach leverages hierarchical priors to navigate a structured taxonomy. This enables the model to effectively generalize by situating a detection between a coarse-grained hypernym (\eg, ``building'') and a more descriptive, fine-grained hyponym (\eg, ``wooden castle'' or ``brick castle''). Leveraging the synergy between these advanced hierarchical perspectives and the long-tailed data characterized in our study, future work may unlock further potential for performance improvements in open-vocabulary environments.

\subsection{Extension to Temporal and Video Domains}
A promising future direction for the MSPL framework lies in its extension to video or temporal tasks. Given its self-contained design, MSPL can be naturally adapted from static images to sequential data. One viable approach involves incorporating a fourth reasoning stage that explicitly accounts for temporal dynamics. In this stage, the model leverages the intermediate evidence accumulated across preceding frames to verify whether bounding boxes with varying appearances correspond to a single persistent object. By aggregating this evidence over multiple timesteps, the framework can generate temporally consistent OV pseudo-labels with refined localization and stable identity (ID) assignments. These labels can subsequently facilitate the training of video detectors through temporal consistency losses or ID stability constraints. Thus, extending the multi-step reasoning process from individual boxes to time-aligned sequences offers a practical and scalable path for video-based OV learning.

\section{Implementation Details}
This section outlines the implementation specificities of our proposed framework, focusing on the integration of the offline pseudo-labeling and online training modules. For a comprehensive description of the baseline architecture and experimental setup, please refer to \cref{appendix:baseline}.

\subsubsection{Offline multi-step pseudo-label generation process.}
During the offline pseudo-labeling phase, we employ SAM--base~\cite{SAM,langsplat} to generate class-agnostic object proposals and utilize Qwen2--7B~\cite{qwen} as the default MLLM for multiple visual reasoning. On the other side, textual embeddings are extracted using a CLIP ViT-B/16~\cite{CLIP} encoder. To enhance representative power, we move beyond simple prompt templates (\eg, ``\texttt{a photo of \{\}}'') used in ViLD~\cite{vild}; instead, we employ a diverse ensemble of hand-crafted templates (\eg, ``\texttt{a scene containing \{\}}'' or ``\texttt{a close-up photo of the \{\}}''). The resulting embeddings are averaged into a single prototypical class embedding. Critically, our pseudo-labels are generated exclusively from the training set without utilizing external image captions and are used solely during the training phase.

\subsubsection{Time efficiency.}
By leveraging an asynchronous producer-consumer architecture, our design allows SAM (\textit{producer}) and CLIP (\textit{consumer}) to process batched data while maximizing VRAM utilization, ensuring a high-throughput and cost-effective pipeline across the dataset. This architectural efficiency enables rapid offline processing for our multi-step pseudo-labeling, even when employing computationally intensive models. As shown in~\cref{tab:pl}, our framework achieves a state-of-the-art amortized labeling latency of 0.43 seconds per image, outperforming existing benchmarks such as PB-OVD~\cite{PB-OVD}, VL-PLM~\cite{VL-PLM}, and SAS-Det~\cite{sasdet}. Following standard evaluation protocols, this metric is calculated by dividing the total labeling duration by the cardinality of the training set on a single A6000 GPU; for instance, the 0.43s latency for OV-COCO is derived as:
\begin{equation}
\text{Latency} = \frac{14 \text{ hours}}{118,287 \text{ images}} \approx 0.43 \text{ seconds/image}.
\end{equation}

In particular, online pseudo-labeling mechanisms like SAS-Det~\cite{sasdet} incur a significant training bottleneck. Specifically, SAS-Det requires approximately 12 hours of self-training on eight A6000 GPUs before the model can function as an effective pseudo-labeler. Despite exhibiting competitive inference latency post-training, the aggregate training overhead of such methods remains substantial. When accounting for the mandatory self-training phase, the amortized per-image latency significantly exceeds the one-second threshold. Consequently, by precluding the requirement for redundant training overhead, our proposed MSPL framework demonstrates superior amortized labeling throughput across the entire training set of OV-COCO.

\subsubsection{Online Contrastive Learning with Multi-Step Supervision.}
We implement MSPL using the Faster R-CNN~\cite{fasterrcnn} framework with a ResNet-50 FPN backbone; for scaling experiments, we utilize a ResNet-50 $\times$4 FPN that is consistent with prior OVD literature. The backbones are initialized with SOCO~\cite{soco} pre-trained weights and trained using synchronized batch normalization~\cite{syncbn}. We follow a $1\times$ training schedule for OV-COCO~\cite{coco} and a $2\times$ schedule for OV-LVIS~\cite{lvis}. Regarding prompt engineering, we employ the hand-crafted templates from ViLD~\cite{vild} for OV-COCO and the learned prompt templates from Detic~\cite{detic} for OV-LVIS. This configuration strictly adheres to the settings established by our baseline to ensure a direct and controlled comparison. For the compositional augmentation of the baseline~\cite{baron}, we mitigate computational overhead by replacing expensive neighborhood region sampling with our semantic anchors, directly bypassing redundant sampling operations, as established in~\cref{sec:online}. 

For RTA and CBL, we construct region and background prototypes using descriptive prompts rather than isolated category names, following the methodology established in ViLD~\cite{vild}. To enhance the robustness of the background prototypes, we utilize a large language model~\cite{gpt4o} to consolidate various background classes within our pseudo-labels into five canonical categories: ``sky,'' ``water surface,'' ``vegetation,'' ``paved ground,'' and ``plain wall.'' Furthermore, we curate a set of object-free templates tailored to these categories (\eg, ``\texttt{clear {} background, no objects}''). These prompts are subsequently tokenized and encoded via a CLIP text encoder, with the resulting embeddings averaged to yield a single, stable representative prototype for each background type.

\section{Hyperparameters}
\label{sec:appx:hyper}
To facilitate a fair comparison, we adopt hyperparameter settings consistent with BARON~\cite{baron}. Specifically, we utilize the SGD optimizer with a momentum of $0.9$ and a weight decay of $2.5 \times 10^{-5}$. The initial learning rate is configured at $0.04$ for OV-COCO and $0.08$ for OV-LVIS. Following standard practice, models are trained for $90,000$ iterations on OV-COCO~\cite{coco} and $180,000$ iterations on OV-LVIS~\cite{lvis}, employing a fixed batch size of $16$ across all experiments. For model selection, checkpoints are recorded every $10,000$ and $30,000$ iterations for COCO and LVIS, respectively, with the best-performing checkpoint on the validation set utilized for final evaluation.

Regarding our proposed modules, we detail the specific configurations for the OV-COCO and OV-LVIS benchmarks. For semantic anchor construction, we filter infrequent pseudo-labels by enforcing a minimum annotation threshold of $1,294$ for OV-COCO and $1$ for OV-LVIS. Furthermore, the temperature parameters for the RTA and CBL objectives are set to $\tau = 0.2$ and $\tau'' = 0.1$, respectively. These parameters modulate the regularization strength of region description and background embeddings relative to their foreground counterparts, ensuring balanced feature alignment across the shared embedding space.

\section{Additional Ablation Study}
In this section, we provide extended experimental analyses to validate the efficacy and comprehensiveness of our proposed framework. Specifically, we conduct a detailed investigation into the impact of visual context modulation (\cref{appendix:preprocessing}) on model performance and present a thorough statistical breakdown of the generated pseudo-label distributions (\cref{sec:sup:plstat}).

\subsection{Visual Context Modulation}
\label{appendix:preprocessing}

\begin{figure}[tb]
  \centering
  \includegraphics[width=\linewidth]{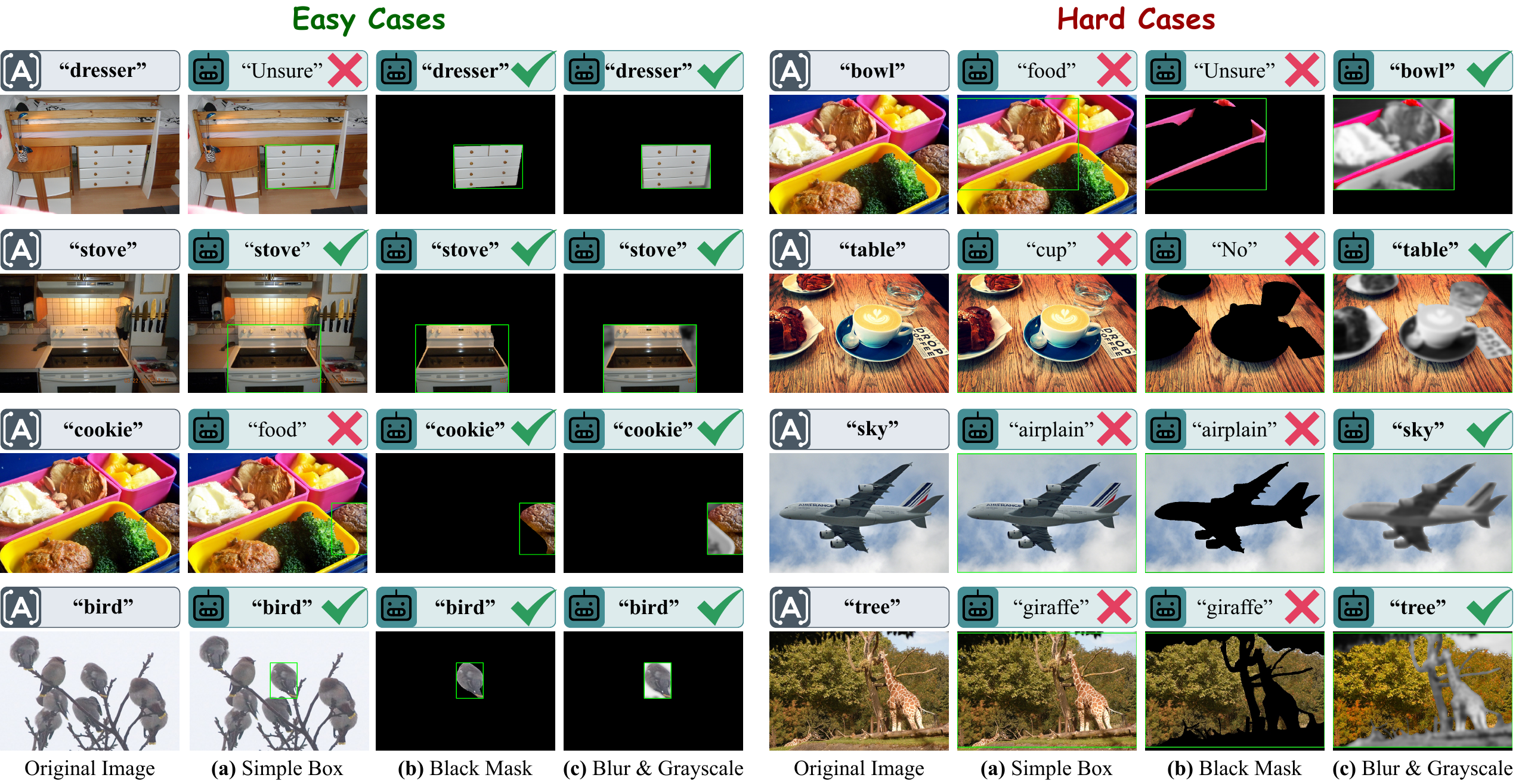}
  \caption{\textbf{Visualization of visual context modulation strategies.} While methods (a) and (b) induce contextual distraction due to background interference, our proposed strategy (c) suppresses extraneous features while maintaining the essential structural context required for robust target reasoning.}
  \label{fig:preprocess}
\end{figure}

We evaluate several image preprocessing strategies aimed at enhancing MLLM focus on specific regions of interest; the quantitative results are summarized in~\cref{tab:preprocessing}, with qualitative visualizations provided in~\cref{fig:preprocess}.

\subsubsection{Simple bounding box.}
We first evaluate the baseline performance using raw image inputs from the OVD benchmark. In this setting, each proposal generated by SAM is demarcated solely by a bounding box prompt (\ie, a green rectangle). This serves as the primary visual cue for the MLLM to localize the target region within the global scene context. Within our experimental framework, where the MLLM is queried on discrete region proposals, empirical results demonstrate that the attenuation of extraneous background noise is critical for maintaining high-fidelity recognition. As evidenced in~\cref{tab:preprocessing}-a, the absence of contextual modulation beyond basic spatial demarcation (\ie, bounding boxes) results in suboptimal performance, yielding an $\text{AP}^{N}_{50}$ of 33.2 (-0.8).

\subsubsection{Black masking.}
A straightforward solution is to apply a binary mask that blacks out all pixels outside the target segmentation boundary. While this focuses the model's attention on the specific region of interest, it eliminates essential contextual cues and can introduce morphological artifacts or silhouette-driven misclassifications~\cite{sepsnet,frankenmask}. In~\cref{fig:preprocess}-b, such artifacts can mislead the model; for instance, a tree may be misclassified as a giraffe due to a coincidentally shaped silhouette. Despite these categorical errors, this masking strategy provides a significant performance boost over raw inputs, achieving 38.7 (+4.7) $\text{AP}^{N}_{50}$.

\subsubsection{Grayscale and Gaussian blurring.}
To mitigate the loss of context inherent in hard masking, we adopt a strategy~\cite{langsplat}, which employs chromatic suppression and spatial smoothing on background regions. Specifically, we apply a BGR-to-grayscale conversion and a Gaussian blur (kernel size $31 \times 31$, $\sigma = 0$) to all areas external to the target mask. This approach maintains the structural context of the scene while effectively de-emphasizing background noise. We validate that this hybrid strategy significantly enhances the reasoning and localization precision of the MLLM---specifically Qwen2~\cite{qwen}---as evidenced in~\cref{tab:preprocessing}-c. Notably, this context-aware mechanism proves effective not only in easy scenarios but also in challenging configurations involving dense spatial overlap. As a result, we adopt this preprocessing pipeline for all region-level queries in our multi-step PL framework, achieving our peak performance of 43.4 $\text{AP}^{N}_{50}$ (+9.4).

\subsection{Analysis of Pseudo-Label Statistics}
\label{sec:sup:plstat}
This section provides a comprehensive statistical overview of our pseudo-labels: categorical distribution and semantic diversity across diverse object categories.

\subsubsection{Class distribution.}
\begin{figure}[tb]
  \centering
  \includegraphics[width=\linewidth]{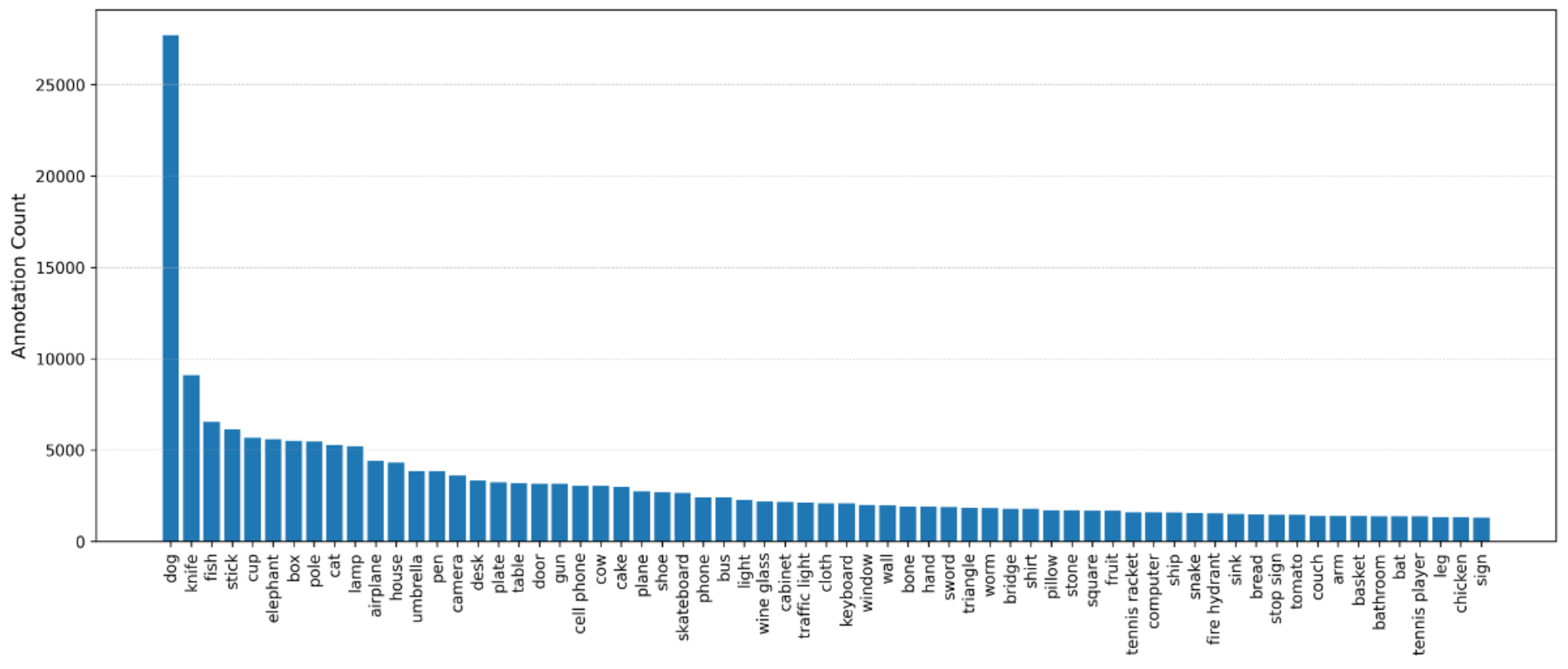}
  \caption{\textbf{Distribution of pseudo-label assignments per category.} Statistics are derived from our Qwen2~\cite{qwen} labeling pipeline across the 65 classes of the OV-COCO benchmark. For visual clarity, we omit the OV-LVIS distribution, which encompasses a significantly larger label space of over 3,000 pseudo-labels.}
  \label{fig:supp:appndx_histogram}
\end{figure}

\Cref{fig:supp:appndx_histogram} illustrates the frequency distribution of pseudo-labels within the OV-COCO dataset, revealing a pronounced long-tailed distribution. A concentrated subset of high-frequency categories accounts for the majority of annotations, reflecting their prevalence and semantic salience within the training corpus. This distribution naturally emerges from the MLLM's propensity to identify commonly occurring objects during the visual reasoning process. Notably, the generated pseudo-labels encompass several novel categories (\eg, ``dog,'' ``knife,'' ``wine glass,'' ``airplane,'' and ``cup''); the inclusion of these labels provides the necessary supervision to significantly improve detection performance on categories absent from the base training set.

\begin{table}[tb]
\centering
\caption{\textbf{Statistical distribution of pseudo-labels aggregated by super-class.} These high-level taxonomic groupings are derived from the generated pseudo-labels with GPT-4o~\cite{gpt4o} on the OV-COCO benchmark.}
\label{tab:stat_super}
\resizebox{\linewidth}{!}{
\begin{tabular}{@{}lccccccccccc@{}}
\toprule
& \rotatebox{45}{Animals} & \rotatebox{45}{Furniture} & \rotatebox{45}{Tools} & \rotatebox{45}{Vehicles} & \rotatebox{45}{Electronics} & \rotatebox{45}{Food} & \rotatebox{45}{Buildings} & \rotatebox{45}{Clothing} & \rotatebox{45}{Shapes} & \rotatebox{45}{Sports} & \rotatebox{45}{Others} \\ \midrule
\textbf{Count} & 11 & 14 & 5 & 6 & 6 & 6 & 3 & 3 & 3 & 3 & 5 \\ \bottomrule
\end{tabular}%
}
\end{table}

\subsubsection{Semantic diversity.}
To gain deeper insight into the semantic composition of our pseudo-labels, we categorize them into broader taxonomic super-classes. This grouping is performed using GPT-4o~\cite{gpt4o}, which is prompted to perform hierarchical clustering, where these specific object categories are grouped into broader, semantically coherent super-classes. As evidenced in~\cref{tab:stat_super}, prominent clusters (\eg, Furniture, Electronics, and Animals) emerge, demonstrating the capability of our pipeline to generate semantically diverse and structurally consistent object categories.

However, we observe that performance degrades for abstract or non-object-level concepts, such as Shapes or Others (\eg, miscellaneous categories), which frequently yield vague or inconsistent reasoning outputs. These findings suggest that while modern MLLMs excel at identifying concrete physical entities, they still struggle with high-level abstractions and functional attributes. This disparity highlights a significant opportunity for future research in enhancing the symbolic reasoning capabilities of vision-language models for OV tasks.

\begin{figure}[tb]
  \centering
  \includegraphics[width=\linewidth]{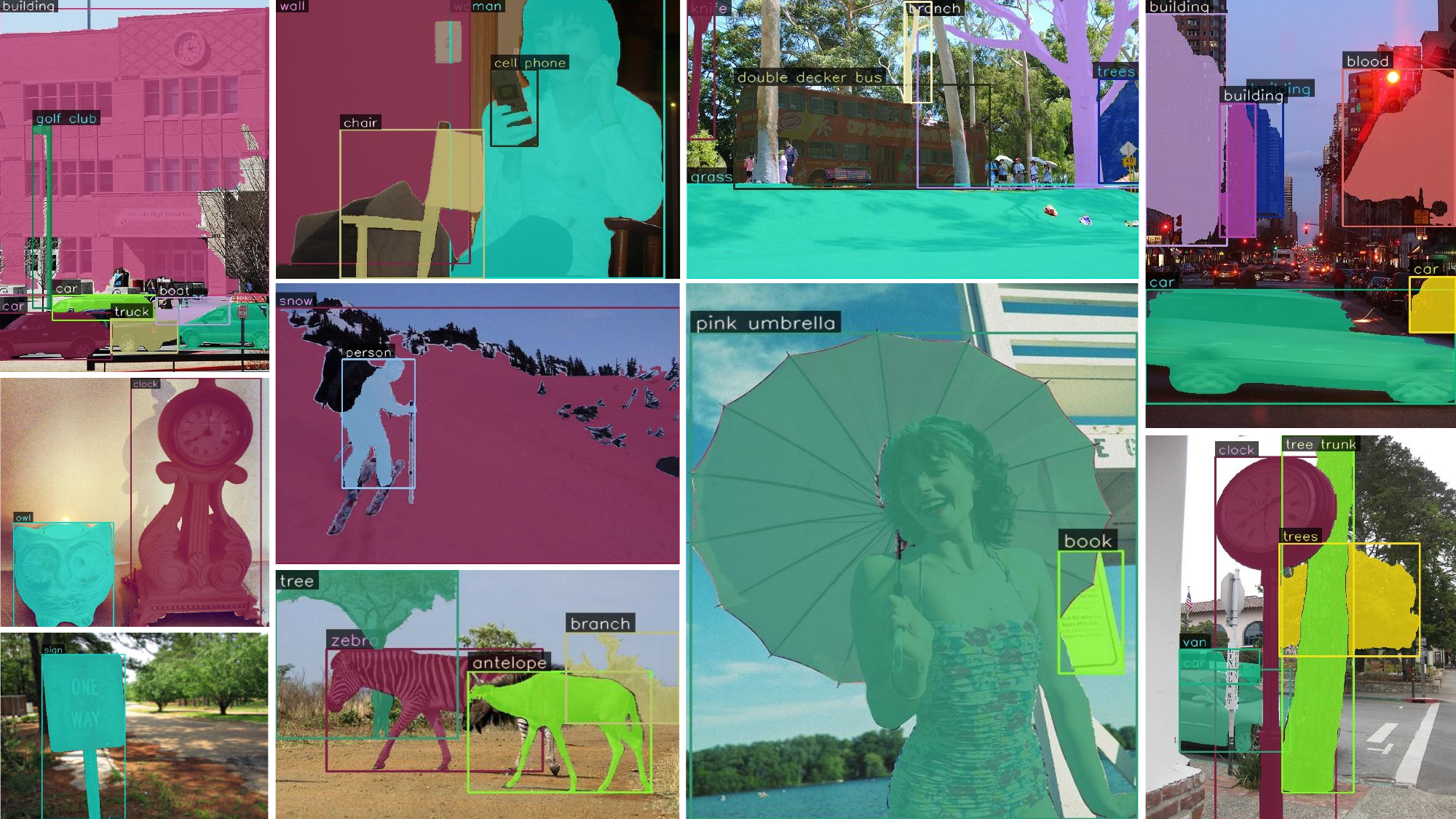}
  \caption{Visualization of our pseudo-annotations on the OV-COCO benchmark.}
  \label{fig:supp:pseudo_vis}
\end{figure}

\section{Additional Qualitative Results}
\label{ablation:qualitative}

\subsection{Qualitative Analysis of Pseudo-Labels}
\subsubsection{Pseudo-label visualization.}
We provide representative visualization examples of our generated pseudo-annotations in \cref{fig:supp:pseudo_vis}, using images sampled from the validation splits of the OVD benchmarks. Each example illustrates the predicted region proposals and their corresponding semantic labels synthesized by our pipeline. These visualizations confirm that our framework generates spatially precise and semantically consistent pseudo-labels across a broad spectrum of object categories. Notably, our method demonstrates robust zero-shot generalization by accurately identifying both frequent base classes and rare novel classes, effectively bridging the supervision gap in open-vocabulary environments.

\subsubsection{Explainability.}
\begin{figure}[tb]
  \centering
  \includegraphics[width=\textwidth, height=\textheight, keepaspectratio]{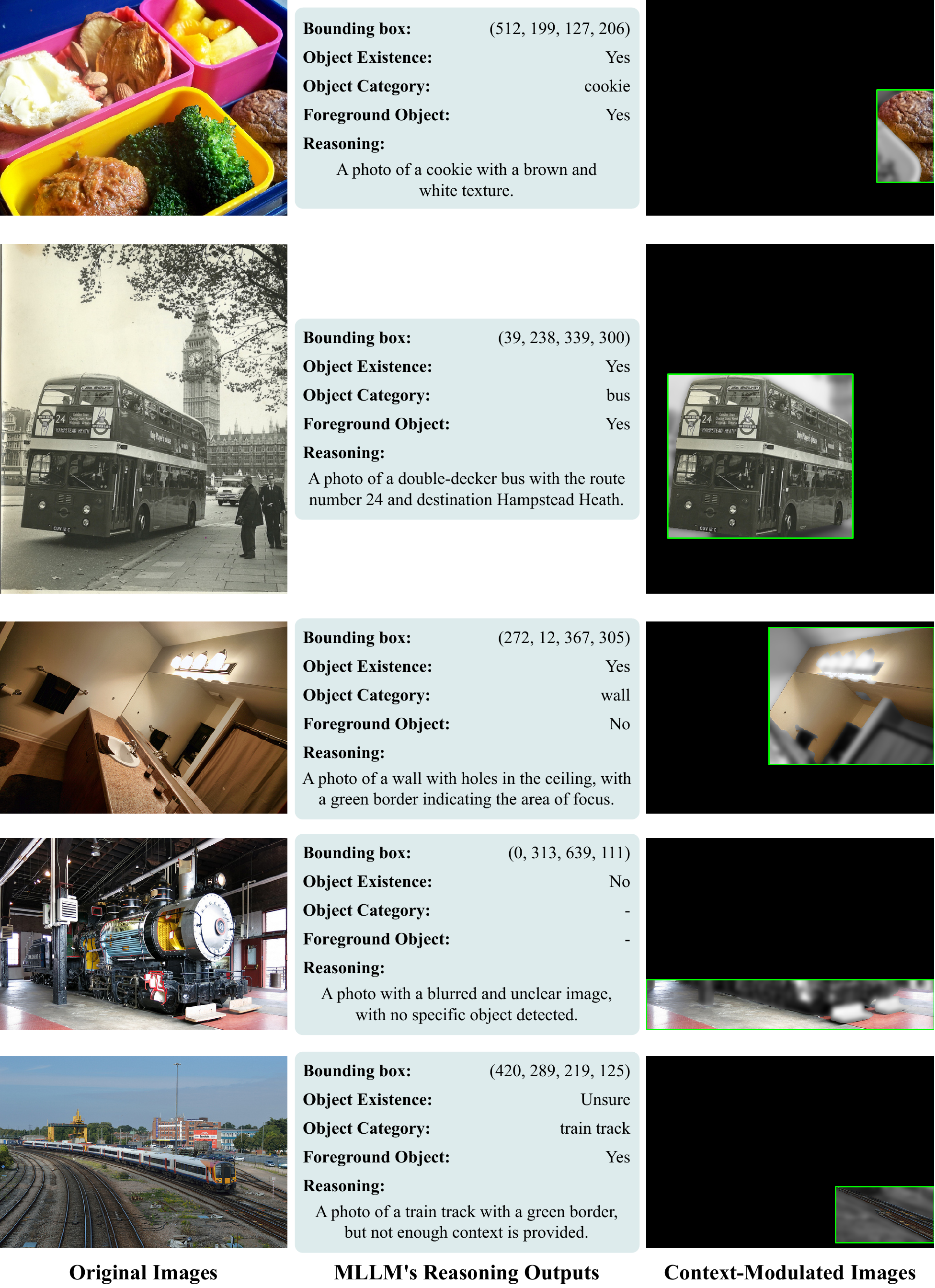}
  \caption{Explainability of our pseudo-annotations on the OV-COCO  benchmark.}
  \label{fig:supp:explain}
\end{figure}

To evaluate the interpretability, we analyze the decision-making logic of MSPL by visualizing the generated labels alongside their reasoning rationales (\cref{fig:supp:explain}), categorizing outcomes into three types: \textit{Affirmative} (``\texttt{Yes}''), \textit{Negative} (``\texttt{No}''), and \textit{Equivocal} (``\texttt{Unsure}'').

\paragraph{Affirmative.}
For high-confidence instances (\eg, ``bus''), MSPL ensures precise spatial grounding and semantic attribution. The multi-step PL mechanism facilitates attribute-level reasoning, identifying defining traits---such as metallic textures or part-whole relationships---before assigning a label. This alignment between visual evidence and textual rationales ensures that pseudo-labels are semantically grounded, robustly enhancing the reliability of the supervision signal.

\paragraph{Negative.}
Ambiguous or non-object regions or artifacts from class-agnostic SAM (\eg, shadows, reflections) are systematically filtered out through our verification stage. In these cases, the reasoning logs typically reveal semantic contradictions, noting a lack of functional attributes or structural integrity to be categorized as a valid object. By pruning spurious proposals via rigorous cross-modal verification, MSPL ensures only semantically valid regions proceed to training, thereby mitigating label noise and preventing over-fitting to hallucinations.

\paragraph{Equivocal.}
Our framework also identifies contextual insufficiency, where a proposal lacks the surrounding cues necessary for definitive identification (\eg, ``train track'' isolated from a train). In such cases, the model often assigns an ``\texttt{Unsure}'' response, facilitating the accumulation of pseudo-labels conducive to robust OVD supervision. This conservative strategy prioritizes high-fidelity supervision over recall, minimizing the propagation of false-positives.

\subsection{Detection Visualization}
\begin{figure}[tb]
  \centering
  \includegraphics[width=\textwidth, height=\textheight, keepaspectratio]{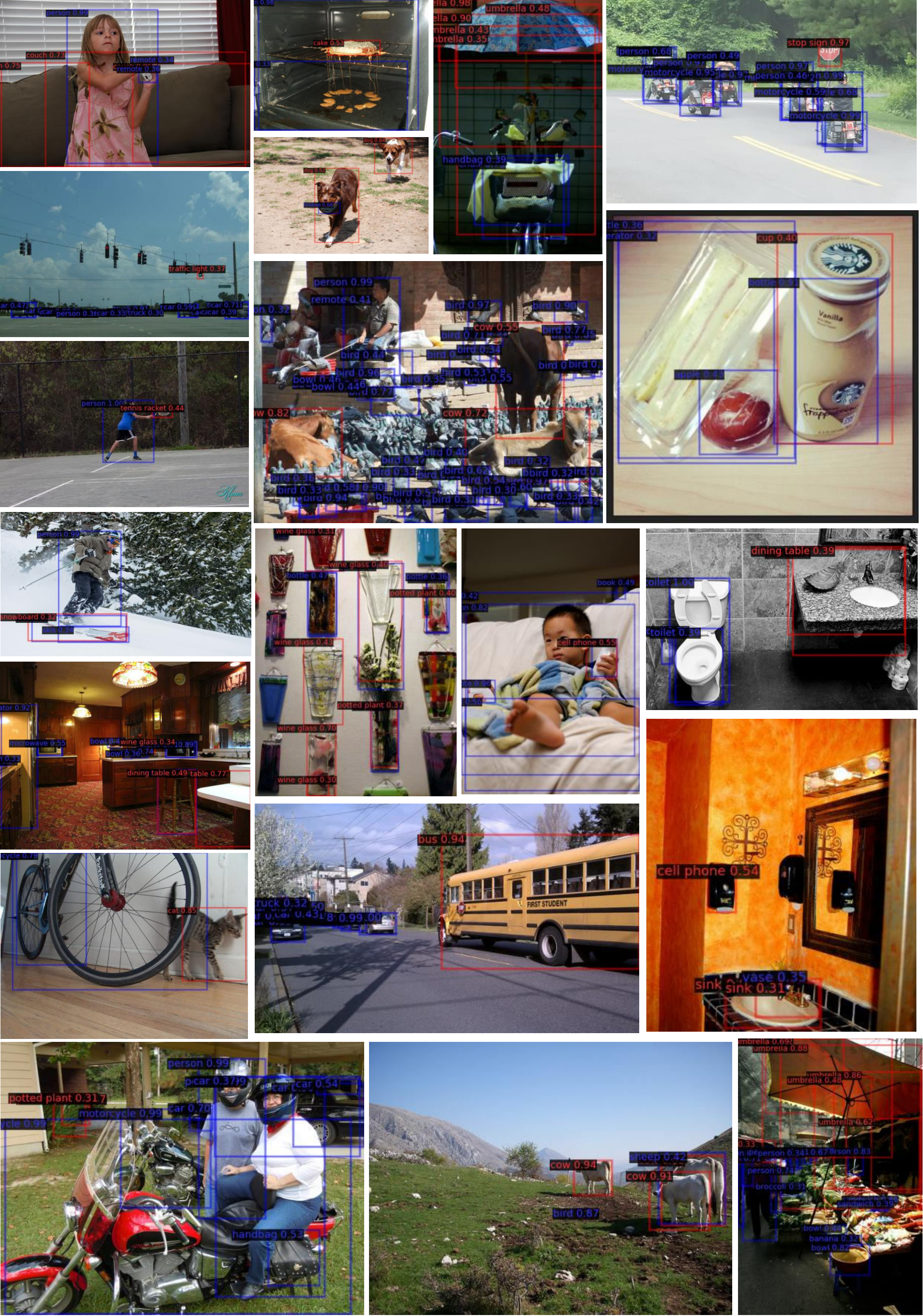}
  \caption{\textbf{Detection results on the OV-COCO dataset.} Novel categories are denoted by red boxes and masks, while base categories are indicated in blue.}
  \label{fig:det_coco}
\end{figure}

\begin{figure}[tb]
  \centering
  \includegraphics[width=\textwidth, height=\textheight, keepaspectratio]{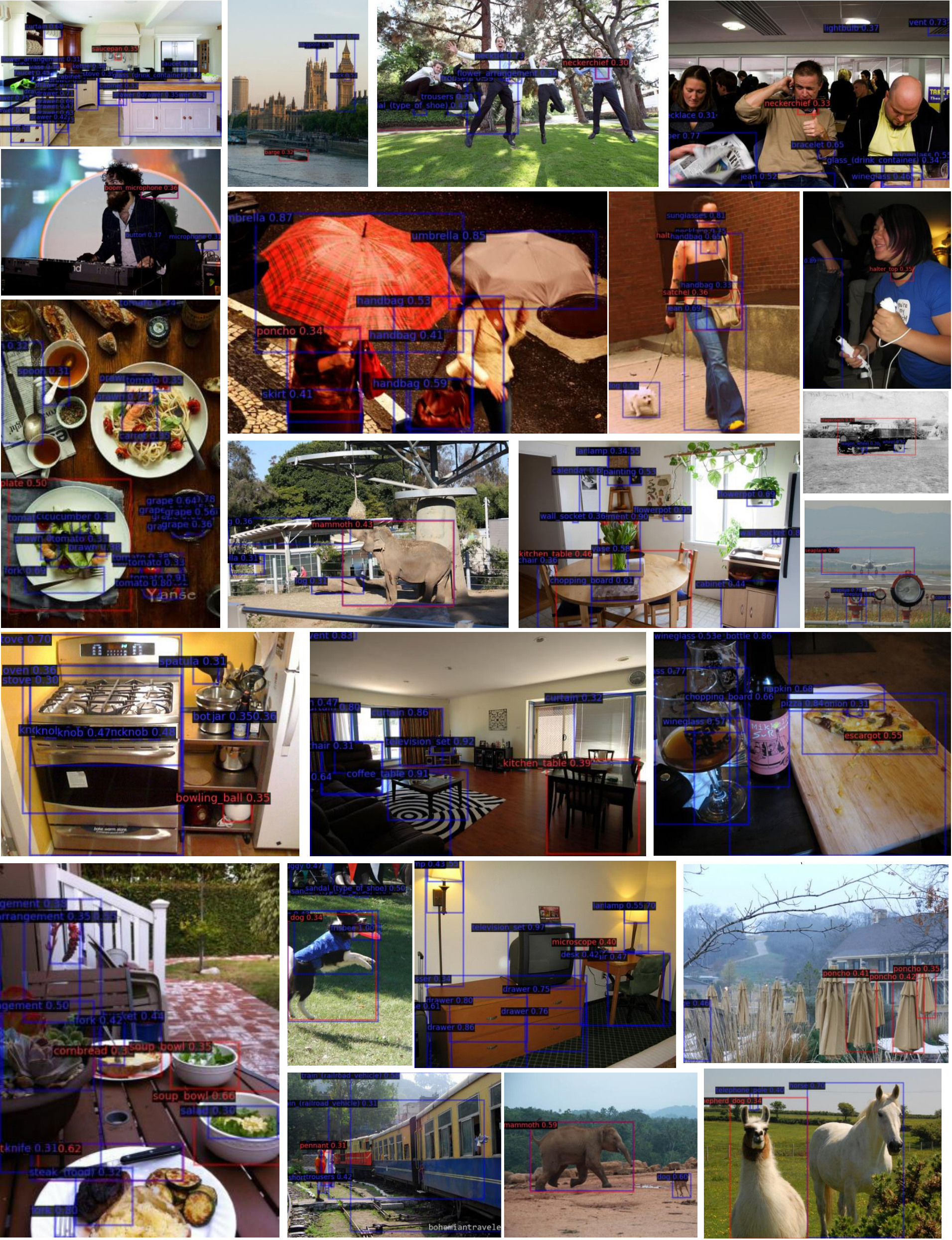}
  \caption{\textbf{Detection results on the OV-LVIS dataset.} Novel (rare) and base categories are distinguished by red and blue bounding boxes and masks, respectively.}
  \label{fig:det_lvis}
\end{figure}

We provide representative detection results of MSPL on OV-COCO~\cite{coco} and OV-LVIS~\cite{lvis} in \cref{fig:det_coco,fig:det_lvis}, respectively. The visualized instances are sampled from the validation splits of each dataset. On OV-COCO, MSPL robustly recognizes novel categories such as ``traffic light,'' ``keyboard,'' and ``snowboard.'' On the more challenging long-tailed OV-LVIS benchmark, it successfully detects rare and fine-grained categories, including ``boom microphone,'' ``mammoth,'' and ``shepherd dog.'' These qualitative results demonstrate MSPL’s ability to generalize across diverse semantic taxonomies and complex scenes.


\clearpage  

\section{Baseline Models}
\label{appendix:baseline}
\subsubsection{Open-vocabulary detectors.}
Recent progress in open-vocabulary detection (OVD) has been catalyzed by the integration of large-scale foundation models, particularly vision-language models (VLMs)~\cite{CLIP, ALIGN}. These models facilitate the recognition of novel categories through strategies such as pseudo-labeling. Our framework extends the Faster R-CNN~\cite{fasterrcnn} architecture by replacing the traditional fixed-category classifier with a linear projection layer that maps regional features into a shared vision-language embedding space. This allows each region to be represented as a set of learned tokens (or pseudo-words''), thereby capturing the nuanced semantics of each object. Given a set of $C$ object categories, the probability $p_c$ that a region belongs to the $c$-th category is computed as:
\begin{equation}
    p_c = \frac{\exp(\tau \cdot \cos(\mathcal{T}(w), f_c))}{\sum_{i=0}^{C-1} \exp(\tau \cdot \cos(\mathcal{T}(w), f_i))},
\end{equation}
where $\mathcal{T}$ denotes the text encoder, $\cos(\cdot, \cdot)$ represents the cosine similarity, and $\tau$ is a temperature scaling factor. Here, $\mathcal{T}(w)$ is the textual embedding of the pseudo-words, and $f_c$ is the category-specific prototype derived from a prompt template (\eg, ``\texttt{a photo of a \{\} in the scene}'').

\paragraph{BARON.} 
Furthermore, we incorporate the core methodology of BARON~\cite{baron} to capture compositional scene structures. During training, the model is optimized using the standard regression and classification losses of Faster R-CNN~\cite{fasterrcnn}, with ground-truth annotations restricted to the base categories. To model contextual relationships, BARON groups neighboring regions for each proposal to form a \textit{bag of regions}. These regions are projected into the word embedding space via a linear layer, yielding a set of \textit{pseudo-words}. 

The bag-of-regions embedding $f_t^i$ is then derived by passing these pseudo-words through the text encoder:
\begin{equation}
    f_t^i = \mathcal{T}(w_0^i + p_0^i, w_1^i + p_1^i, \dots, w_{N^i-1}^i + p_{N^i-1}^i),
\end{equation}
where $N^i$ denotes the number of regions in the $i$-th bag, and $p_j^i$ represents the positional embedding for the $j$-th region. Subsequently, $f_t^i$ is aligned with the corresponding VLM image embedding $f_v^i = \mathcal{V}(b_0^i, b_1^i, \dots, b_{N_i^i})$, where $b_j^i$ corresponds to the $j$-th regional visual feature. To enforce this alignment, BARON employs a bidirectional InfoNCE-based~\cite{infonce} contrastive loss:
\begin{equation}
    \mathcal{L}_{bag} = -\frac{1}{2G} \sum_{k=0}^{G-1} \left( \log (p_{t,v}^k) + \log (p_{v,t}^k) \right),
\end{equation}
where the conditional probabilities are defined as:
\begin{equation}
    p_{t,v}^k = \frac{\exp(\tau' \cdot \langle f_t^k, f_v^k \rangle)}{\sum_{i=0}^{G-1} \exp(\tau' \cdot \langle f_t^k, f_v^i \rangle)}, \quad
    p_{v,t}^k = \frac{\exp(\tau' \cdot \langle f_v^k, f_t^k \rangle)}{\sum_{i=0}^{G-1} \exp(\tau' \cdot \langle f_v^k, f_t^i \rangle)}.
\end{equation}
Here, $G$ represents the total number of bags and $\tau'$ is a temperature scaling factor. This mechanism enables the model to leverage the rich compositional structures inherent in pre-trained VLMs.

To maintain fine-grained alignment for individual regions, BARON additionally utilizes an individual-level contrastive loss:
\begin{equation}
    \mathcal{L}_{\text{individual}} = -\frac{1}{2N} \sum_{k=0}^{N-1} \left( \log (q_{t,v}^k) + \log (q_{v,t}^k) \right),
\end{equation}
with the region-specific probabilities given by:
\begin{equation}
    q_{t,v}^k = \frac{\exp(\tau_{\text{ind}} \cdot \langle g_t^k, g_v^k \rangle)}{\sum_{i=0}^{N-1} \exp(\tau_{\text{ind}} \cdot \langle g_t^k, g_v^i \rangle)}, \quad
    q_{v,t}^k = \frac{\exp(\tau_{\text{ind}} \cdot \langle g_v^k, g_t^k \rangle)}{\sum_{i=0}^{N-1} \exp(\tau_{\text{ind}} \cdot \langle g_v^k, g_t^i \rangle)}.
\end{equation}
In this formulation, $N$ is the total number of regions, $g_t^k$ and $g_v^k$ denote the teacher and student embeddings for the $k$-th region, respectively, and $\tau_{\text{ind}}$ is the temperature parameter for similarity re-scaling.

\begin{figure}[tb]
  \centering
  \includegraphics[width=\linewidth]{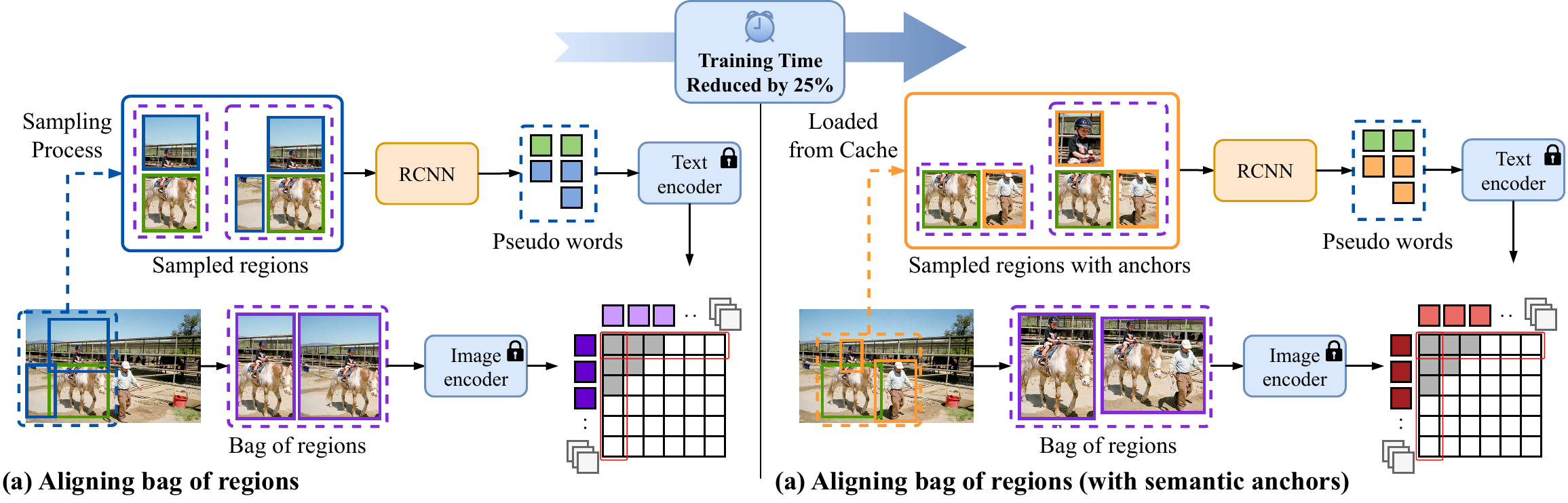}
  \caption{\textbf{BARON architecture~\cite{baron} with our semantic anchor-accelerated compositional augmentation.} Our proposed anchor-caching strategy serves as an efficient alternative to BARON's naive sampling. This optimization accelerates training by 1.5$\times$ relative to the baseline without any degradation in detection accuracy.}
  \label{fig:baseline}
\end{figure}

\subsubsection{Segment Anything Model (SAM).}
SAM~\cite{SAM} is a foundational segmentation framework designed to predict fine-grained instance masks based on spatial prompts (\eg, points or boxes). By enabling zero-shot segmentation, SAM provides class-agnostic object candidates essential for downstream OV perception tasks~\cite{ovsam,boostingsam}. Building upon this, recent work such as LangSplat~\cite{langsplat} leverages SAM’s dense point sampling to extract hierarchical masks across various spatial granularities---subpart, part, and whole levels---derived from confidence scores and spatial containment criteria. This structured decomposition facilitates precise feature extraction and underpins the construction of language-aware 3D representations by effectively capturing the compositional nature of visual entities. In addition, this multi-scale strategy allows the framework to resolve semantic queries at different levels of abstraction, from part components to entire object instances, significantly improving the flexibility of OV scene understanding.

\subsubsection{Multimodel LLMs (MLLMs).}
MLLMs have emerged as a powerful paradigm for OVD, as they bridge the gap between low-level visual perception and high-level semantic reasoning through instruction following~\cite{led,ccktdet,mllmsurvey,gvt,llmdet}.

\paragraph{BLIP-2.}
BLIP-2~\cite{blip2} introduces a modular framework designed to decouple visual feature extraction from linguistic reasoning. It consists of a frozen image encoder, a frozen LLM (\eg, OPT~\cite{opt}), and a trainable Querying Transformer (Q-Former)~\cite{qformer} that serves as a lightweight bottleneck for vision-language alignment. This architecture facilitates efficient cross-modal representational learning, achieving competitive performance on benchmarks such as image captioning and visual question answering with minimal computational overhead.

\paragraph{InstructBLIP.}
InstructBLIP~\cite{instructblip} extends the BLIP-2 architecture through large-scale instruction tuning. By integrating a ViT-G vision encoder~\cite{vit} with a frozen Flan-T5~\cite{flant5} backbone, the model is optimized to follow complex natural language instructions across a wide array of multimodal tasks. As summarized in~\cref{tab:stat_mllm}, InstructBLIP demonstrates robust zero-shot generalization on academic benchmarks, maintaining an accuracy range of 24\%--32\% on diverse reasoning tasks.

\paragraph{Qwen2-VL.}
Qwen2-VL~\cite{qwen} represents a state-of-the-art series of multilingual MLLMs, with parameter scales ranging from 0.5B to 72B. Trained on high-quality, web-scale multimodal corpora, it utilizes an optimized tokenizer and positional encoding (M-ROPE) to enhance multilingual and spatial understanding. Qwen2-VL exhibits superior reasoning and instruction-following capabilities, notably achieving an 81.0\% score on MMBench~\cite{mmbench}. This high degree of vision-language alignment makes it an ideal candidate for the complex multi-step reasoning required in our pseudo-labeling pipeline.

\begin{table}[tb]
\centering
\caption{Evaluation of MLLM zero-shot capabilities on multimodal benchmarks.}
\resizebox{\textwidth}{!}{%
\begin{tabular}{l c c c c}
\toprule
\textbf{Model} & \textbf{MMBench V1.1}~\cite{mmbench} & \textbf{MMStar}~\cite{mmstar} & \textbf{MMMU}~\cite{mmmu} & \textbf{HallusionBench Avg.}~\cite{hallubench} \\
\midrule
BLIP2 (2.7B)~\cite{blip2}           & -   & -    & -    & - \\
InstructBLIP-7B~\cite{instructblip} & 28.4& 32.7 & 30.6 & 31.2 \\
Qwen2-VL-7B~\cite{qwen}             & 81.0& 60.7 & 53.7 & 50.4 \\
\bottomrule
\end{tabular}%
}
\label{tab:stat_mllm}
\end{table}

\section{Prompts}
\label{appendix:prompt}
\begin{figure}[tb]
  \centering
  \includegraphics[width=\textwidth, height=\textheight, keepaspectratio]{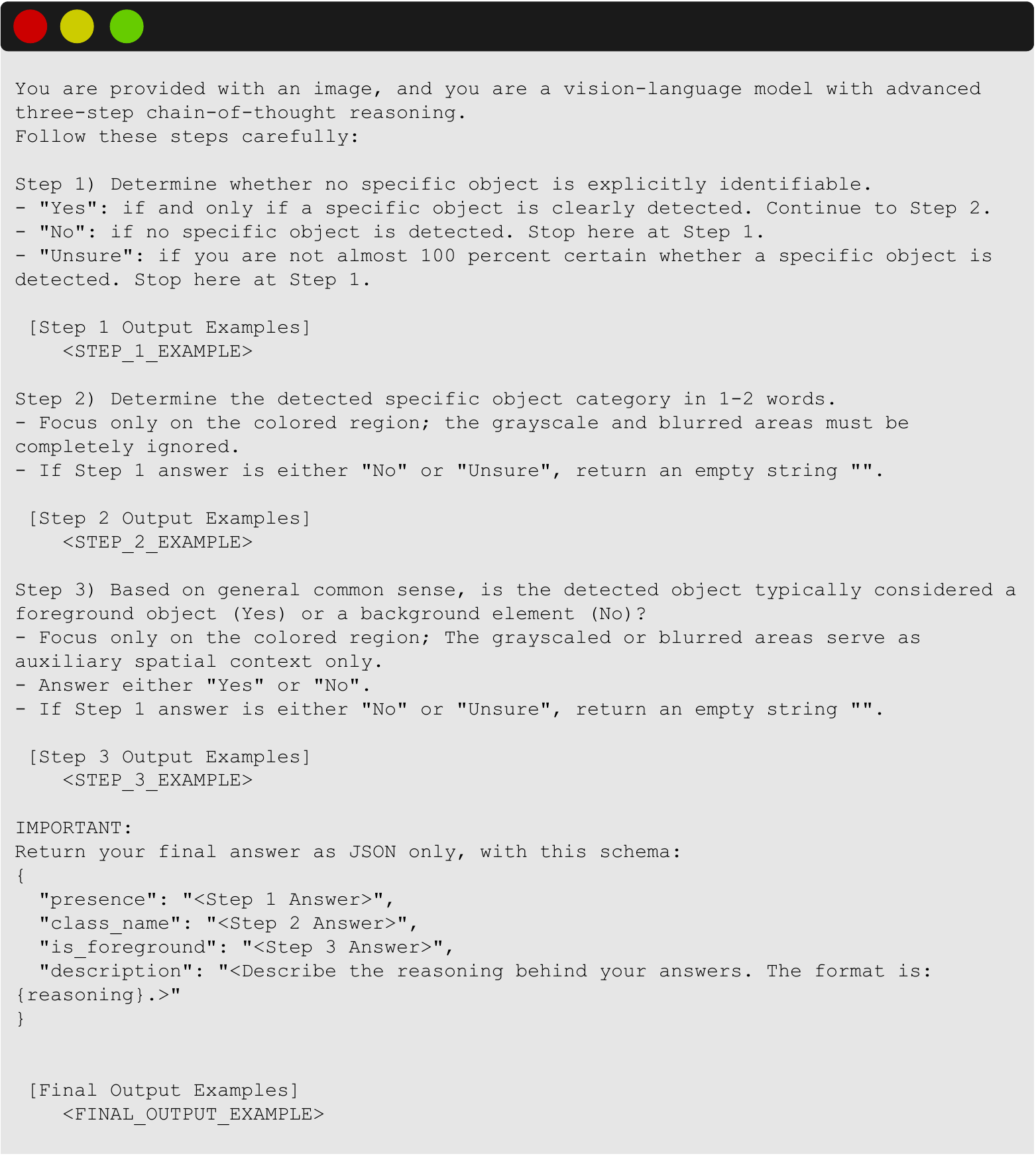}
  \caption{Prompt template for the three-stage pseudo-labeling pipeline.}
  \label{fig:prompt}
\end{figure}

In this section, we detail the specialized prompt architecture developed for our offline multi-step pseudo-labeling pipeline. As detailed in~\cref{fig:prompt}, this structured prompting strategy is designed to elicit incremental reasoning steps from the MLLM, ensuring that each object classification is supported by explicit visual and semantic evidence. The prompt follows a multi-stage template that directs the model to evaluate spatial coordinates, object attributes, and foreground-background relationships before arriving at a final class assignment.

\end{document}